\documentclass{article}

\usepackage[preprint]{neurips_2026}
\workshoptitle{WORKSHOP TITLE}

\usepackage[utf8]{inputenc}
\usepackage[T1]{fontenc}
\usepackage{hyperref}
\usepackage{url}
\usepackage{booktabs}
\usepackage{longtable}
\newcommand{\dashrule}[1]{\multicolumn{#1}{@{}c@{}}{\leavevmode\cleaders\hbox to 2.5pt{\hss\vrule height 0.4pt depth 0pt width 1pt\hss}\hskip 0pt plus 1filll\kern0pt}\\}
\usepackage{multirow}
\usepackage{amsfonts}
\usepackage{amsmath}
\usepackage{nicefrac}
\usepackage{microtype}
\usepackage[table,dvipsnames]{xcolor}
\usepackage{graphicx}
\usepackage{wrapfig}
\usepackage{float}
\usepackage{enumitem}

\title{Turning TS2Vec into a Foundation Model\\
       for Time Series Classification}

\author{%
  Vasilii Feofanov\\
  42 NL\\
  \texttt{vfeofanov@42.com} 
  \And Zhenghan Tai \\ University of Toronto \AND Andrei Ermakov \\ 42 NL \And Irina Rish \\ Université de Montréal \\ Mila - Quebec AI Institute
}

\begin{document}

\maketitle

\begin{abstract}
    TS2Vec is one of the strongest self-supervised methods for time series classification, introduced in 2022. It is designed to be trained anew on every dataset: self-supervised pre-training followed by linear probing. Since then, the field has shifted towards foundation models that reuse one encoder across datasets. Yet TS2Vec offers a compact convolutional architecture that operates directly on temporal samples and accepts variable-length inputs without resampling. We ask whether these properties can be retained when the encoder is pre-trained only once, and show that they can: TS2Vec can essentially be turned into a foundation model. Three changes suffice: pre-train on a synthetic corpus, feed the encoder a raw \emph{amplitude} view and an instance-normalized \emph{shape} view of the input through two small stems, and distill one specialist encoder per view into a single encoder. The result, TS2Vec-FM, is a 1.6M-parameter dilated convolutional network that, kept frozen, reaches 83.3\% on UCR-128 and 75.3\% on UEA-27 under linear probing, within 0.3 pp of MantisV2, the current state of the art, and above TS2Vec trained separately on each target dataset. What we find most interesting is how different TS2Vec-FM is from other foundation models in design: near-parity is reached without attention, without a patch tokenizer, and without intermediate-layer extraction, using an ``old-school'' convolutional encoder and a two-layer MLP stem.
\end{abstract}
  
\section{Introduction}
\label{sec:intro}

Time series foundation models (TSFMs) aim to provide a single frozen encoder whose
features are useful across downstream datasets, removing the need to train a
specialized model per task~\citep{feofanov2026mantis,lin2024nutime,goswami2024moment}. 
Most of TSFMs use transformer architecture pre-trained on a large corpus of time series, real and / or synthetic \citep{xie2026cauker}.
Prior to foundation models, the same problem was attacked from another angle: self-supervised
methods such as TS2Vec~\citep{yue2022ts2vec}, T-Loss~\citep{franceschi2019tloss} and
TNC~\citep{tonekaboni2021unsupervised} train a small encoder \emph{separately on each
target dataset}. TS2Vec in particular remains a standard and strong baseline, but its
design is viewed as somewhat outdated, and it has never been evaluated as a
universal encoder itself. This paper asks whether it can become one.

\begin{figure}[t]
  \centering
  \includegraphics[width=\linewidth]{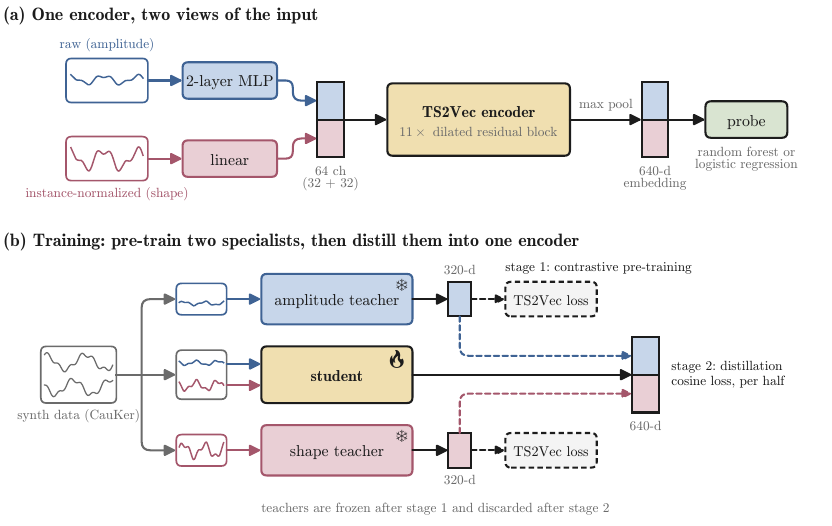}
  \caption{TS2Vec-FM architecture and training. (a) Raw and instance-normalized inputs pass through an MLP and a linear stem, respectively. Their concatenation feeds a shared convolutional trunk and temporal max pooling. (b) Two specialists are pretrained on synthetic data with the TS2Vec objective and then frozen. On the same data, each teacher supervises one 320-dimensional half of the student through timestamp-level and pooled cosine losses. Downstream, only a classifier on frozen student features is fitted.}
  \label{fig:teaser}
\end{figure}

The answer is yes, with three changes (Figure~\ref{fig:teaser}): pretrain once on synthetic data;
give the encoder raw and instance-normalized input views through separate stems; and
distill~\citep{hinton2015distilling} two resulting specialists into a single encoder.
The specialists retain the original TS2Vec architecture and contrastive objective,
while the student uses the same residual-block design with modified input and output layers and is trained by distillation.

The result is notable for two reasons. First, the resulting 1.6M-parameter
encoder is competitive: frozen, it outperforms per-dataset TS2Vec and the
evaluated forecasting-based extractors, and comes within a few tenths of a
percentage point of MantisV2 while using about 25\% fewer parameters.
Second, it reaches this performance with a markedly different design ---
convolutions instead of attention, a two-layer MLP stem instead of a
hand-designed multi-view tokenizer, and the final representation rather than a
selected intermediate layer. This near-parity suggests that the information
exposed to the encoder may matter more than the specific tokenizer design;
our ablations point in particular to the complementarity of amplitude and shape.

\section{From per-dataset TS2Vec to a foundation model}
\label{sec:method}

\textbf{Original TS2Vec.} The reference encoder first maps each timestamp to
64 channels, then applies ten hidden residual blocks and one output block with
exponentially increasing dilation, producing a 320-dimensional representation
at every timestamp. It is trained on aligned regions of two overlapping
contextual crops with a hierarchical objective that contrasts both timestamps
within a series and instances within a batch. A global series embedding is the
temporal maximum of these representations. Appendix~\ref{app:ts2vec} specifies
the architecture, crops, and losses exactly.

\textbf{Step 1: the pre-training regime.} We pretrain once on univariate synthetic series generated by CauKer~\citep{xie2026cauker} algorithm, so all encoder weight updates use synthetic data.
The advantage of the TS2Vec architercture is that a convolutional encoder needs no fixed input size,
so series keep their native lengths. However, similar to most time series foundation models,
our encoder is univariate, so for multivariate datasets, we encode channels independently 
with the embeddings concatenated.

\textbf{Step 2: normalization.} Original TS2Vec does not apply instance-wise normalization.
For classification, however, absolute signal values may contain discriminative information. We therefore pretrain separate TS2Vec encoders on raw and instance-normalized inputs.
The results (Table \ref{tab:recipe}) have been quite unexpected. The encoder without normalization 
performs significantly better on the multivariate (UEA) datasets, while the encoder with normalization greatly outperforms
its counterpart on the univariate (UCR) datasets: instance normalization gains 2.9 pp on UCR but costs 1.5 pp on UEA,
and the raw-input encoder is the mirror image.
We suggest that this phenomenon is explained by the nature of contrastive pre-training which looks for shortcuts to
discriminate different samples in the embedding space. 
In the un-normalized case, the shortcut is the amplitude of the signal, so embeddings encode mostly this kind of information. 
In the normalized case, the shortcut is rather the shape of the signal. 
More arguments to this hypothesis gives the fact that in many multivariate datasets a deviation of one channel 
compared to the others gives a discriminative signal, so encoding signal amplitude is important.

\textbf{Step 3: architecture.} Given the aforementioned results, we slightly modify the encoder to include both normalized and raw views: 
the instance-normalized view is linearly mapped into 32 channels, while the raw view is embedded 
through a two-layer MLP with a GELU nonlinearity. 
Their concatenation forms the 64-channel input of the shared convolutional part. So, we retain the TS2Vec residual-block design and increase the output dimension from 320 to 640, yielding a 1.64M-parameter encoder.
We note that the asymmetry between the two views matters as the raw view needs the extra nonlinearity to align its encodings with the normalized one.

\textbf{Step 4: distillation.} 
% Direct contrastive pre-training of the dual-input
% encoder is numerically fragile at the batch-512, $4\times10^{-3}$ learning-rate
% setting used for throughput: two nominally identical 1M runs differed by 4.15 pp
% because gradient explosions occurred at different steps. Clipping the global
% gradient norm at 10 reduces the repeat gap to at most 0.15 pp over the first three epochs and yields 80.52\% UCR accuracy at both 1M and 2M. 
The performance discrepancy between the normalized and un-normalized encoders has motivated us to use them as teachers (further also called specialists) to distill their knowledge into our dual-input single encoder. We use a factorized objective, matching each half of the 640-dimensional student representation to its corresponding teacher by cosine loss,
per timestamp and after pooling. 
What is interesting is that directly pre-training the dual-input encoder with the usual TS2Vec contrastive objective yields inferior results (Table \ref{tab:recipe}). 
% The direct contrastive term is then unnecessary, so we drop it: the student reaches 81.1\% at the 100K scale in a single
% encoder, against 80.5\% for the contrastively trained dual trunk and 81.5\% for the two-specialist concatenation.

\section{Experimental Results}
\label{sec:experiments}

\textbf{Setup.}
Further, we refer our approach (synthetic pre-training, dual-input architecture, distillation) as \textbf{TS2Vec-FM}. We compare it with other methods by evaluating frozen representations on UCR-128~\citep{dau2019ucr} and UEA-27~\citep{bagnall2018uea},
following \cite{feofanov2026mantis}.
Dataset-specific classifiers are a 200-tree random forest (RF), averaged over three predictor seeds,
and logistic regression (LogReg) preceded by StandardScaler.
Pre-training checkpoints are selected by UCR RF accuracy at seed 42; UEA is not used for checkpoint selection.
We compare against Catch22+~\citep{lubba2019catch22}, MOMENT~\citep{goswami2024moment},
TiRex~\citep{auer2025tirex}, Chronos-2~\citep{ansari2025chronos2},
TiViT-H / TiConvNext~\citep{roschmann2025tivit}, NuTime~\citep{lin2024nutime},
and Mantis (V2 checkpoint, \citealp{feofanov2026mantis,feofanov2026mantisv2closingzeroshotgap}).

\begin{table}[t]
  \caption{Zero-shot feature extraction accuracy (\%), with parameters counted after layer
  pruning where applicable. The dashed rule separates the
  original per-dataset TS2Vec, which is not a foundation model: it trains one encoder
  on each dataset's own training split (128 UCR and 27 UEA encoders of 0.6M
  parameters). $^\ddagger$With its original RBF-SVM probe it reaches 82.10 on UCR.}
  \label{tab:main}
  \centering
  \begin{tabular}{llcccccc}
    \toprule
    & & & \multicolumn{2}{c}{Random forest} & \multicolumn{2}{c}{LogReg} \\
    \cmidrule(lr){4-5}\cmidrule(lr){6-7}
    Model & Type & Params & UCR & UEA & UCR & UEA \\
    \midrule
    Catch22+ & statistics & --- & 79.69 & 69.63 & 77.04 & 67.56 \\
    MOMENT & forecasting & 161.4M & 77.89 & 69.91 & 79.84 & 71.74 \\
    NuTime & classification & 2.0M & 77.32 & 71.99 & 79.05 & 74.91 \\
    TiRex & forecasting & 16.5M & 80.13 & 69.52 & 82.02 & 72.48 \\
    Chronos-2 & forecasting & 44.0M & 80.02 & 69.67 & 82.44 & 72.43 \\
    TiViT-H & vision & 276.6M & 79.43 & 70.91 & 82.88 & 73.30 \\
    TiConvNext & vision & 180.3M & 80.29 & 71.05 & 83.47 & 73.50 \\
    MantisV2 & classification & 2.2M & \textbf{81.95} & \textbf{74.20} & \textbf{83.60} & \textbf{75.52} \\
    \midrule
    % TS2Vec-FM, two specialists & classification & 1.27M & 81.70 & 74.03 & \textbf{83.66} & 75.36 \\
    TS2Vec-FM, one encoder & classification & 1.64M & 81.76 & 73.60 & 83.32 & 75.30 \\
    \noalign{\vskip -0.2cm}
    \dashrule{7}
    \noalign{\vskip -0.0cm}
    \multicolumn{7}{l}{\emph{Reference (not zero-shot: one encoder trained per target dataset)}} \\
    \noalign{\vskip 0.07cm}
    TS2Vec, per dataset$^\ddagger$ & self-supervised & 0.6M each & 79.83 & 69.50 & 81.36 & 70.79 \\
    \bottomrule
  \end{tabular}
\end{table}

\textbf{The recipe produces a competitive TSFM} (Table~\ref{tab:main}).
The proposed TS2Vec-FM shows the second-best result, trailing MantisV2 by 0.19 / 0.60 pp with RF and 0.28 / 0.22 pp with LogReg,
respectively, while using about 25\% fewer parameters. At the same time, it outperforms the forecasting- and vision-based extractors
(10--170$\times$ as many parameters), except TiConvNext on UCR LogReg. We would like to note that unlike MantisV2, our encoder uses native lengths and does not need interpolate each series to a fixed length. In Appendix~\ref{app:scaling}, we show that interpolating series worsen the performance of TS2Vec-FM.

\textbf{It also beats TS2Vec trained per dataset.} Under both the random forest and logistic regression probe, the margin between the foundation model and per-dataset encoder is more that 1.9 \% on UCR. The gap widens even more on multivariate UEA-27, even though TS2Vec-FM was pre-trained on univariate data. Note that in the original TS2Vec paper, they use an RBF-SVM with hyperparameters tuned, which increases the probing performance, though still below TS2Vec-FM with logistic regression. We have also tried to reproduce the results published by \cite{yue2022ts2vec} in two different setups: first, our reimplementation of the original protocol that reproduces the published UCR-128 performance to within 0.8 pp with a per-dataset correlation of 0.984, and second, we have run the
authors' unmodified code that reproduces fifteen of the sixteen evaluable headline
metrics of the original paper (across all classification, forecasting and anomaly
detection) within 5\% relative (see Appendix~\ref{app:perdataset}).

\begin{table}[t]
  \caption{What each ingredient contributes (random forest, \%) at 100K, 1M and 2M
  synthetic series. Instance normalization trades UEA for UCR at every scale, the
  specialists and the direct contrastive baseline do not improve with more data, and
  distillation consolidates the two specialists into the stronger single encoder at
  every scale. $^\dagger$mean over three pre-training seeds; $^\ddagger$with gradient
  clipping.}
  \label{tab:recipe}
  \centering
  \small
  \setlength{\tabcolsep}{4pt}
  \begin{tabular}{llcccccc}
    \toprule
    & & \multicolumn{2}{c}{100K series} & \multicolumn{2}{c}{1M series} & \multicolumn{2}{c}{2M series} \\
    \cmidrule(lr){3-4}\cmidrule(lr){5-6}\cmidrule(lr){7-8}
    Encoder input & Training & UCR & UEA & UCR & UEA & UCR & UEA \\
    \midrule
    raw only & contrastive & 77.43$^\dagger$ & 72.02$^\dagger$ & 76.99 & 71.56 & 77.68 & 72.22 \\
    normalized only & contrastive & 80.35$^\dagger$ & 70.56$^\dagger$ & 80.26 & 69.78 & 79.89 & 70.19 \\
    % both, two separate encoders & contrastive, concatenated & 81.53$^\dagger$ & 73.94$^\dagger$ & -- & -- & 81.70 & 74.03 \\
    dual-input & contrastive & 80.52$^\dagger$ & 73.41$^\dagger$ & 80.52$^\ddagger$ & 73.41$^\ddagger$ & 80.52$^\ddagger$ & 73.07$^\ddagger$ \\
    dual-input & distillation (ours) & \textbf{81.08}$^\dagger$ & \textbf{73.90}$^\dagger$ & \textbf{81.62}$^\dagger$ & \textbf{73.48}$^\dagger$ & \textbf{81.76} & \textbf{73.60} \\
    \bottomrule
  \end{tabular}
\end{table}

\textbf{Ablation of the recipe} (Table~\ref{tab:recipe}). We first ask whether more synthetic data helps. For the ingredients taken alone the answer is negative: from 100K to 1M and 2M series, the two specialists and the dual-input encoder pre-trained directly with the contrastive TS2Vec loss stay within seed noise or slightly degrade (the instance-normalized specialist loses 0.5 pp on UCR). The distilled encoder is the only one that benefits: on UCR it gains 0.54 pp from 100K to 1M (81.08 to 81.62\%, three pre-training seeds each, seed standard deviation 0.12 pp) and a further 0.14 pp at 2M, while on UEA it stays flat (73.9, 73.5 and 73.6\%). We detail the scaling experiments further in Appendix~\ref{app:scaling}.
The comparison with the directly pre-trained dual-input encoder also isolates the contribution of distillation: it is superior at every scale, and its margin widens with data, from 0.56 pp at 100K to 1.10 pp at 1M and 1.24 pp at 2M on UCR. We also note that we had to use gradient clipping for the contrastive baseline due to numerical instability. The pattern suggests that direct training underuses the complementary views, while
already-specialized teachers keep both subspaces alive. This is, in effect, a
learned substitute for what hand-designed TSFM tokenizers provide by construction when
they devote separate branches to the normalized signal and to raw patch
statistics~\citep{feofanov2026mantis,lin2024nutime}.

\section{Discussion}
\label{sec:discussion}

In this paper, we introduced TS2Vec-FM, showing that synthetic pretraining and specialist distillation can make a compact
convolutional encoder competitive for cross-dataset classification while retaining
native-length processing.
The results motivate revisiting earlier self-supervised architectures, adapting them to the current setup and thereby enriching the family of available time series foundation models.

\textbf{Limitations.}
The 2M headline model is trained at a single pre-training seed, although the
100K ablations and the 1M student use three seeds.
TS2Vec-FM remains slightly below MantisV2 on all four metrics, so our claim is
competitive performance with a smaller encoder rather than superiority.
Evaluation is limited to UCR and UEA, and whether specialist distillation
generalizes to other backbones and tokenizer designs remains open.

\bibliographystyle{plainnat}
\bibliography{references}

@inproceedings{ijcai2021-324,
  title     = {Time-Series Representation Learning via Temporal and Contextual Contrasting},
  author    = {Eldele, Emadeldeen and Ragab, Mohamed and Chen, Zhenghua and Wu, Min and Kwoh, Chee Keong and Li, Xiaoli and Guan, Cuntai},
  booktitle = {Proceedings of the Thirtieth International Joint Conference on Artificial Intelligence, {IJCAI-21}},
  pages     = {2352--2359},
  year      = {2021},
}

@article{tonekaboni2021unsupervised,
  title={Unsupervised representation learning for time series with temporal neighborhood coding},
  author={Tonekaboni, Sana and Eytan, Danny and Goldenberg, Anna},
  journal={arXiv preprint arXiv:2106.00750},
  year={2021}
}

@article{yue2022ts2vec, 
  title={{TS2Vec}: Towards Universal Representation of Time Series}, 
  volume={36}, 
  url={https://ojs.aaai.org/index.php/AAAI/article/view/20881}, 
  DOI={10.1609/aaai.v36i8.20881}, 
  number={8}, 
  journal={Proceedings of the AAAI Conference on Artificial Intelligence}, 
  author={Yue, Zhihan and Wang, Yujing and Duan, Juanyong and Yang, Tianmeng and Huang, Congrui and Tong, Yunhai and Xu, Bixiong}, 
  year={2022}, 
  month={Jun.}, 
  pages={8980-8987} 
}

@article{goswami2024moment,
  title={{MOMENT}: A family of open time-series foundation models},
  author={Goswami, Mononito and Szafer, Konrad and Choudhry, Arjun and Cai, Yifu and Li, Shuo and Dubrawski, Artur},
  journal={arXiv preprint arXiv:2402.03885},
  year={2024}
}

@article{lin2024nutime,
title={{NuTime}: Numerically Multi-Scaled Embedding for Large- Scale Time-Series Pretraining},
author={Chenguo Lin and Xumeng Wen and Wei Cao and Congrui Huang and Jiang Bian and Stephen Lin and Zhirong Wu},
journal={Transactions on Machine Learning Research},
issn={2835-8856},
year={2024},
url={https://openreview.net/forum?id=TwiSBZ0p9u},
note={}
}

@article{dau2019ucr,
  title={The {UCR} time series archive},
  author={Dau, Hoang Anh and Bagnall, Anthony and Kamgar, Kaveh and Yeh, Chin-Chia Michael and Zhu, Yan and Gharghabi, Shaghayegh and Ratanamahatana, Chotirat Ann and Keogh, Eamonn},
  journal={IEEE/CAA Journal of Automatica Sinica},
  volume={6},
  number={6},
  pages={1293--1305},
  year={2019},
  publisher={IEEE}
}

@article{bagnall2018uea,
  title={The {UEA} multivariate time series classification archive, 2018},
  author={Bagnall, Anthony and Dau, Hoang Anh and Lines, Jason and Flynn, Michael and Large, James and Bostrom, Aaron and Southam, Paul and Keogh, Eamonn},
  journal={arXiv preprint arXiv:1811.00075},
  year={2018}
}

@inproceedings{
xie2026cauker,
title={CauKer: Classification Time Series Foundation Models Can Be Pretrained on Synthetic Data},
author={Shifeng Xie and Vasilii Feofanov and Jianfeng Zhang and Themis Palpanas and Ievgen Redko},
booktitle={The Fourteenth International Conference on Learning Representations},
year={2026},
url={https://openreview.net/forum?id=xBW2FIfswU}
}

@article{roschmann2025tivit,
      title={Time Series Representations for Classification Lie Hidden in Pretrained Vision Transformers}, 
      author={Roschmann, Simon and Bouniot, Quentin and Feofanov, Vasilii and Redko, Ievgen and Akata, Zeynep},
      year={2025},
      journal={arXiv preprint arXiv:2506.08641},
      url={https://arxiv.org/abs/2506.08641}, 
}

@article{auer2025tirex,
  title={{TiRex}: Zero-Shot Forecasting Across Long and Short Horizons with Enhanced In-Context Learning},
  author={Auer, Andreas and Podest, Patrick and Klotz, Daniel and B{\"o}ck, Sebastian and Klambauer, G{\"u}nter and Hochreiter, Sepp},
  journal={arXiv preprint arXiv:2505.23719},
  year={2025}
}

@article{auer2025tirexclassification,
  title={Pre-trained Forecasting Models: Strong Zero-Shot Feature Extractors for Time Series Classification},
  author={Auer, Andreas and Klotz, Daniel and B{\"o}ck, Sebastinan and Hochreiter, Sepp},
  journal={arXiv preprint arXiv:2510.26777},
  year={2025}
}

@article{lubba2019catch22,
  title={catch22: {CA}nonical {T}ime-series {CH}aracteristics: Selected through highly comparative time-series analysis},
  author={Lubba, Carl H and Sethi, Sarab S and Knaute, Philip and Schultz, Simon R and Fulcher, Ben D and Jones, Nick S},
  journal={Data mining and knowledge discovery},
  volume={33},
  number={6},
  pages={1821--1852},
  year={2019},
  publisher={Springer}
}

@inproceedings{feofanov2026mantis,
title={Mantis: Lightweight Foundation Model for Time Series Classification},
author={Vasilii Feofanov and Songkang Wen and Shifeng Xie and Simon Roschmann and Marius Alonso and Hongbo Guo and Romain Ilbert and Malik Tiomoko and Quentin Bouniot and Zeynep Akata and Lujia Pan and Jianfeng Zhang and Ievgen Redko},
booktitle={Forty-third International Conference on Machine Learning},
year={2026},
url={https://openreview.net/forum?id=gbJMAjXLZ4}
}

@article{ansari2025chronos2,
  title={Chronos-2: From Univariate to Universal Forecasting},
  author={Ansari, Abdul Fatir and Shchur, Oleksandr and K{\"u}ken, Jaris and Auer, Andreas and Han, Boran and Mercado, Pedro and Rangapuram, Syama Sundar and Shen, Huibin and Stella, Lorenzo and Zhang, Xiyuan and others},
  journal={arXiv preprint arXiv:2510.15821},
  year={2025}
}

@article{franceschi2019tloss,
  title={Unsupervised scalable representation learning for multivariate time series},
  author={Franceschi, Jean-Yves and Dieuleveut, Aymeric and Jaggi, Martin},
  journal={Advances in neural information processing systems},
  volume={32},
  year={2019}
}

@inproceedings{zerveas2021transformer,
  title={A transformer-based framework for multivariate time series representation learning},
  author={Zerveas, George and Jayaraman, Srideepika and Patel, Dhaval and Bhamidipaty, Anuradha and Eickhoff, Carsten},
  booktitle={Proceedings of the 27th ACM SIGKDD conference on knowledge discovery \& data mining},
  pages={2114--2124},
  year={2021}
}

@article{moroshan2025tempopfn,
  title={TempoPFN: Synthetic Pre-training of Linear RNNs for Zero-shot Time Series Forecasting},
  author={Moroshan, Vladyslav and Siems, Julien and Zela, Arber and Carstensen, Timur and Hutter, Frank},
  journal={arXiv preprint arXiv:2510.25502},
  year={2025}
}

@inproceedings{hinton2015distilling,
  title={Distilling the Knowledge in a Neural Network},
  author={Hinton, Geoffrey and Vinyals, Oriol and Dean, Jeff},
  booktitle={NeurIPS Deep Learning and Representation Learning Workshop},
  year={2015}
}

@misc{feofanov2026mantisv2closingzeroshotgap,
      title={MantisV2: Closing the Zero-Shot Gap in Time Series Classification with Synthetic Data and Test-Time Strategies}, 
      author={Vasilii Feofanov and Songkang Wen and Jianfeng Zhang and Lujia Pan and Ievgen Redko},
      year={2026},
      eprint={2602.17868},
      archivePrefix={arXiv},
      primaryClass={cs.LG},
      url={https://arxiv.org/abs/2602.17868}, 
}

@misc{moakher2026utica,
      title={UTICA: Multi-Objective Self-Distllation Foundation Model Pretraining for Time Series Classification}, 
      author={Yessin Moakher and Youssef Attia El Hili and Vasilii Feofanov},
      year={2026},
      eprint={2603.01348},
      archivePrefix={arXiv},
      primaryClass={cs.LG},
      url={https://arxiv.org/abs/2603.01348}, 
}

@inproceedings{lee2024softclt,
  title={Soft contrastive learning for time series},
  author={Lee, Seunghan and Park, Taeyoung and Lee, Kibok},
  booktitle={International Conference on Learning Representations},
  volume={2024},
  pages={46815--46839},
  year={2024}
}

@inproceedings{luo2023infots,
  title={Time series contrastive learning with information-aware augmentations},
  author={Luo, Dongsheng and Cheng, Wei and Wang, Yingheng and Xu, Dongkuan and Ni, Jingchao and Yu, Wenchao and Zhang, Xuchao and Liu, Yanchi and Chen, Yuncong and Chen, Haifeng and others},
  booktitle={Proceedings of the AAAI conference on artificial intelligence},
  volume={37},
  number={4},
  pages={4534--4542},
  year={2023}
}

@inproceedings{liu2024timesurl,
  title={Timesurl: Self-supervised contrastive learning for universal time series representation learning},
  author={Liu, Jiexi and Chen, Songcan},
  booktitle={Proceedings of the AAAI conference on artificial intelligence},
  volume={38},
  number={12},
  pages={13918--13926},
  year={2024}
}

@article{foumani2024series2vec,
  title={Series2vec: similarity-based self-supervised representation learning for time series classification: NM Foumani et al.},
  author={Foumani, Navid Mohammadi and Tan, Chang Wei and Webb, Geoffrey I and Rezatofighi, Hamid and Salehi, Mahsa},
  journal={Data Mining and Knowledge Discovery},
  volume={38},
  number={4},
  pages={2520--2544},
  year={2024},
  publisher={Springer}
}

@misc{comad2025,
      title={CoMAD: A Multiple-Teacher Self-Supervised Distillation Framework}, 
      author={Sriram Mandalika and Lalitha V},
      year={2025},
      eprint={2508.04816},
      archivePrefix={arXiv},
      primaryClass={cs.CV},
      url={https://arxiv.org/abs/2508.04816}, 
}

@misc{orourke2026causaldag,
      title={A Causal DAG Prior for Synthetic Time-Series Classification Datasets}, 
      author={Franco Martino O'Rourke and Ana Trisovic and Dimitris Bertsimas},
      year={2026},
      eprint={2606.21776},
      archivePrefix={arXiv},
      primaryClass={cs.LG},
      url={https://arxiv.org/abs/2606.21776}, 
}

@misc{nagpal2026mix,
      title={Mix, Don't Pick: Why Synthetic Corpus Composition Matters for Time Series Foundation Model Pretraining}, 
      author={Aaryan Nagpal and Debdeep Sanyal and Murari Mandal and Dhruv Kumar and Saurabh Deshpande},
      year={2026},
      eprint={2606.09912},
      archivePrefix={arXiv},
      primaryClass={cs.LG},
      url={https://arxiv.org/abs/2606.09912}, 
}

\appendix

\section{TS2Vec: architecture and objective}
\label{app:ts2vec}

This section gives more details on the TS2Vec model and training objective. Our implementation follows the authors' released
code~\citep{yue2022ts2vec}.
% ; the modifications introduced by our method begin
% only at the input stems and the distillation objective described in
% Section~\ref{sec:method}.

\begin{figure}[H]
  \centering
  \includegraphics[width=\linewidth]{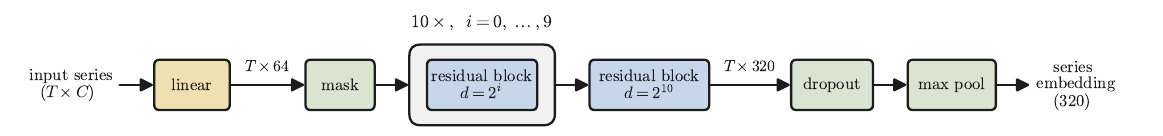}
  \caption{Reference TS2Vec encoder. Ten width-64 residual blocks with dilation
  $2^{i}$ are followed by a width-320 output block; one block is detailed in
  Figure~\ref{fig:ts2vec-block}.}
  \label{fig:ts2vec-backbone}
\end{figure}

\begin{wrapfigure}[24]{r}{0.27\linewidth}
  \vspace{-1.2\baselineskip}
  \centering
  \includegraphics[width=0.85\linewidth]{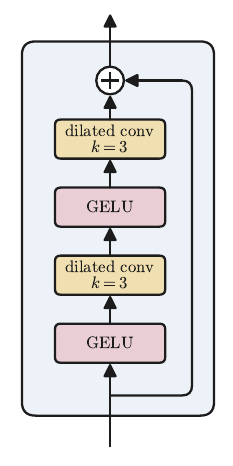}
  \caption{One TS2Vec residual block: two GELU--dilated-convolution pairs
  plus a skip connection (identity, or a $1\times1$ convolution when the
  width changes).}
  \label{fig:ts2vec-block}
  \vspace{-\baselineskip}
\end{wrapfigure}
\paragraph{Encoder.}
Let $\mathbf{X}_i=(\mathbf{x}_{i,1},\ldots,\mathbf{x}_{i,T})
\in\mathbb{R}^{T\times C}$ denote series $i$. A learned affine map sends each
timestamp from $C$ input channels to a hidden width $H=64$, with
$\mathbf{W}_{\mathrm{in}}\in\mathbb{R}^{H\times C}$ and
$\mathbf{b}_{\mathrm{in}}\in\mathbb{R}^{H}$. During pre-training,
an independent mask $m_{i,t}\sim\operatorname{Bernoulli}(0.5)$ is sampled for
each contextual view, and a masked timestamp is replaced by the zero vector
after the affine map:
\begin{equation}
  \mathbf{h}^{(0)}_{i,t}
  =m_{i,t}\bigl(\mathbf{W}_{\mathrm{in}}\mathbf{x}_{i,t}
  +\mathbf{b}_{\mathrm{in}}\bigr).
  \label{eq:ts2vec-input}
\end{equation}
Missing timestamps are masked in the same way. The trunk contains eleven
residual blocks indexed by $k=0,\ldots,10$, each with two same-padded
one-dimensional convolutions of kernel size three and dilation
$\delta_k=2^k$. Writing $\mathcal{C}_{k,1}$ and $\mathcal{C}_{k,2}$ for these
convolutions, a block is
\begin{align}
  \mathbf{u}^{(k)} &=
    \mathcal{C}_{k,1}\!\left(\operatorname{GELU}
    (\mathbf{h}^{(k)})\right), \\
  \mathbf{h}^{(k+1)} &=
    \mathcal{C}_{k,2}\!\left(\operatorname{GELU}
    (\mathbf{u}^{(k)})\right)
    +\mathcal{P}_k(\mathbf{h}^{(k)}).
  \label{eq:ts2vec-block}
\end{align}
The first ten blocks preserve width 64; the final block maps 64 to the
representation dimension $D=320$. The residual projection $\mathcal{P}_k$ is
the identity when dimensions agree and a pointwise convolution otherwise.
Dropout with probability 0.1 follows the trunk. The output of the last block is
the timestamp representation
$\mathbf{r}_{i,t}=\mathbf{h}^{(11)}_{i,t}\in\mathbb{R}^{D}$; we drop the series
index $i$ when it is not needed. For a univariate input this encoder contains
637,248 trainable parameters.

\paragraph{Input length and receptive field.}
The encoder is fully convolutional and the readout is a max-pool over all
timestamps, so the model accepts any length $T$ with cost linear in $T$, and the
series embedding always depends on every timestamp. What is bounded is the
context of each timestamp representation: with symmetric padding, $\mathbf{r}_t$
sees at most $1+4\sum_{k=0}^{10}2^k=8189$ timestamps ($\pm 4094$ around $t$).
For $T\le 8189$ every $\mathbf{r}_t$ observes the whole series; this holds for
all 128 UCR datasets (longest: Rock, $T=2844$) and for 26 of the 27 UEA datasets
(MotorImagery, $T=3000$). Beyond that bound no single $\mathbf{r}_t$ sees the
whole series and the only cross-window integration is the order-free max-pool,
so a class that is defined by relating two events more than $\sim$8k steps apart
cannot be represented, whereas locally detectable patterns still are. The one
benchmark case beyond the bound, EigenWorms ($T=17{,}984$), is nevertheless
handled well (88.0\% RF and 90.8\% LogReg versus 81.4\% and 88.5\% for MantisV2,
Appendix~\ref{app:additional}). A second, more practical limitation is the
\emph{trained} receptive field: pre-training used 512-timestamp series, which
the 8189-timestamp field covers many times over, so the long-range taps of the
last blocks (dilations 128 to 1024) were mostly fitted against zero padding
and only receive signal at both ends on long inputs. We do not quantify this
shift; downsampling very long inputs or pre-training on longer series are the
natural remedies.

\paragraph{Contextual crops.}
For each minibatch, TS2Vec samples an overlap length
$L$ uniformly from $\{2,\ldots,T\}$. It samples an overlap $[a,b)$ with $b-a=L$, a left extension
boundary $a_{\mathrm{e}}\in\{0,\ldots,a\}$, and an extended right boundary
$b_{\mathrm{e}}\in\{b,\ldots,T\}$. Each series receives a shift
$o_i\in\{-a_{\mathrm{e}},\ldots,T-b_{\mathrm{e}}\}$. The two encoder inputs are
\begin{equation}
  \widetilde{\mathbf{X}}^{\,A}_i
    =\mathbf{X}_i[o_i+a_{\mathrm{e}}:o_i+b],
  \qquad
  \widetilde{\mathbf{X}}^{\,B}_i
    =\mathbf{X}_i[o_i+a:o_i+b_{\mathrm{e}}].
  \label{eq:ts2vec-crops}
\end{equation}
Their contexts differ, but the last $L$ outputs of the first view and the first
$L$ outputs of the second refer to the same timestamps. Stacking these aligned
timestamp representations $\mathbf{r}_{i,t}$ over the minibatch gives
$\mathbf{A},\mathbf{B}\in\mathbb{R}^{B\times L\times D}$, with minibatch size
$B$.

\paragraph{Dual contrastive loss.}
For an anchor $\mathbf{q}$, positive $\mathbf{p}$ and negative set
$\mathcal{N}$, define
\begin{equation}
  \ell(\mathbf{q},\mathbf{p};\mathcal{N})
  =-\log
  \frac{\exp(\mathbf{q}^{\top}\mathbf{p})}
       {\exp(\mathbf{q}^{\top}\mathbf{p})
        +\sum_{\mathbf{n}\in\mathcal{N}}
         \exp(\mathbf{q}^{\top}\mathbf{n})}.
  \label{eq:ts2vec-nce}
\end{equation}
The implementation uses raw dot products: it applies neither $\ell_2$
normalization nor a temperature parameter. The negative collections below are
indexed lists rather than mathematical sets: equal-valued vectors remain
distinct denominator terms, and $\mathbin{\Vert}$ denotes list concatenation.
Instance contrast fixes a timestamp $t$ and treats the other series in both
views as negatives:
\begin{align}
  \mathcal{N}^{A}_{\mathrm{inst}}(i,t)
    &=(\mathbf{A}_{j,t})_{j\ne i}
      \mathbin{\Vert}(\mathbf{B}_{j,t})_{j\ne i},\\
  \mathcal{N}^{B}_{\mathrm{inst}}(i,t)
    &=(\mathbf{B}_{j,t})_{j\ne i}
      \mathbin{\Vert}(\mathbf{A}_{j,t})_{j\ne i},\\
  \mathcal{L}_{\mathrm{inst}}(\mathbf{A},\mathbf{B})
    &=\frac{1}{2BL}\sum_{i=1}^{B}\sum_{t=1}^{L}
      \Bigl[
      \ell(\mathbf{A}_{i,t},\mathbf{B}_{i,t};
           \mathcal{N}^{A}_{\mathrm{inst}}(i,t))
      +\ell(\mathbf{B}_{i,t},\mathbf{A}_{i,t};
           \mathcal{N}^{B}_{\mathrm{inst}}(i,t))
      \Bigr].
  \label{eq:ts2vec-instance}
\end{align}
Temporal contrast instead fixes a series $i$ and treats all other aligned
timestamps in both views as negatives. Thus
\begin{align}
  \mathcal{N}^{A}_{\mathrm{temp}}(i,t)
    &=(\mathbf{A}_{i,s})_{s\ne t}
      \mathbin{\Vert}(\mathbf{B}_{i,s})_{s\ne t},\\
  \mathcal{N}^{B}_{\mathrm{temp}}(i,t)
    &=(\mathbf{B}_{i,s})_{s\ne t}
      \mathbin{\Vert}(\mathbf{A}_{i,s})_{s\ne t},\\
  \mathcal{L}_{\mathrm{temp}}(\mathbf{A},\mathbf{B})
    &=\frac{1}{2BL}\sum_{i=1}^{B}\sum_{t=1}^{L}
      \Bigl[
      \ell(\mathbf{A}_{i,t},\mathbf{B}_{i,t};
           \mathcal{N}^{A}_{\mathrm{temp}}(i,t))
      +\ell(\mathbf{B}_{i,t},\mathbf{A}_{i,t};
           \mathcal{N}^{B}_{\mathrm{temp}}(i,t))
      \Bigr].
  \label{eq:ts2vec-temporal}
\end{align}
Both objectives are symmetric across the two views. The positive is included
once in the denominator of Equation~\eqref{eq:ts2vec-nce}; only the anchor
self-similarity is excluded.

\paragraph{Hierarchical contrasting.}
Set $(\mathbf{A}^{(0)},\mathbf{B}^{(0)})=(\mathbf{A},\mathbf{B})$ and obtain
scale $s+1$ by max-pooling each representation over adjacent pairs of
timestamps with stride two:
$L_{s+1}=\lfloor L_s/2\rfloor$. For odd $L_s$, the unmatched final timestamp
is dropped, exactly as in the released implementation. This continues until
$L_{S-1}=1$, where
$S=\lfloor\log_2 L\rfloor+1$. Define
$\beta_s=\mathbf{1}[L_s>1]$, which switches temporal contrast off at the
singleton scale. The optimized objective is
\begin{equation}
  \begin{aligned}
  \mathcal{L}_{\mathrm{hier}}
  =\frac{1}{S}\sum_{s=0}^{S-1}\Big[
    &\alpha\,\mathcal{L}_{\mathrm{inst}}
      (\mathbf{A}^{(s)},\mathbf{B}^{(s)})\\
    &+\beta_s(1-\alpha)\,\mathcal{L}_{\mathrm{temp}}
      (\mathbf{A}^{(s)},\mathbf{B}^{(s)})\Big],
  \qquad \alpha=\tfrac{1}{2}.
  \end{aligned}
  \label{eq:ts2vec-hierarchical}
\end{equation}
Consequently, fine scales discriminate timestamps as well as series, while the
singleton top scale contributes instance contrast only. The original paper
writes the unweighted sum
$\mathcal{L}_{\mathrm{inst}}+\mathcal{L}_{\mathrm{temp}}$ at each non-singleton
scale; the released code uses $\alpha=\tfrac12$ as above, which differs by the
global factor $\tfrac12$. Equation~\eqref{eq:ts2vec-hierarchical} states the
implementation used in our experiments.

\paragraph{Optimization and readout.}
We use AdamW with learning rate $10^{-3}$, weight decay $10^{-2}$ and batch size
8 when reproducing per-dataset TS2Vec. The foundation-model specialists use
batch size 128 at the same learning rate; the throughput experiments use batch
size 512 and learning rate $4\times10^{-3}$. Checkpoints contain a running
arithmetic average initialized from the starting weights and updated after every
optimizer step. At test time masking and dropout are disabled, timestamp
representations are max-pooled over the complete series. Original per-dataset
TS2Vec maps all $C$ channels jointly through $\mathbf{W}_{\mathrm{in}}$ and
therefore always returns one 320-dimensional vector. Our foundation encoder is
pre-trained on univariate synthetic series; when evaluating it on multivariate
data, we instead encode the $C$ channels independently and concatenate them,
giving a $320C$-dimensional frozen representation.

\section{Related work}
\label{sec:related}
\textbf{Self-supervised representation learning for time series.}
A first line of work learns representations without labels by training an encoder on the
target dataset and then fitting a simple classifier on its frozen output. T-Loss uses a
triplet loss over subsequences~\citep{franceschi2019tloss}, TNC contrasts temporal
neighborhoods~\citep{tonekaboni2021unsupervised}, TS-TCC combines temporal and contextual
contrasting~\citep{ijcai2021-324}, TST reconstructs masked timestamps with a
transformer~\citep{zerveas2021transformer}, and TS2Vec optimizes a hierarchical contrastive
loss over two overlapping views of each series~\citep{yue2022ts2vec}. TS2Vec has remained a
standard baseline, and subsequent work refines its objective through soft contrastive
assignments~\citep{lee2024softclt}, information-aware augmentation~\citep{luo2023infots},
frequency-domain augmentation~\citep{liu2024timesurl}, and augmentation-free similarity
prediction~\citep{foumani2024series2vec}. What these methods share is the training regime
rather than the objective: the encoder is fitted to the dataset on which it is evaluated.
Transfer between datasets has been examined within this line, but from a single real source
dataset and in comparison with other per-dataset
methods~\citep{franceschi2019tloss,yue2022ts2vec}.

\textbf{Foundation models for time series classification.}
A second line pre-trains one encoder on a large corpus, keeps it frozen, and fits only a
probe on each downstream dataset. NuTime~\citep{lin2024nutime},
MOMENT~\citep{goswami2024moment}, Mantis~\citep{feofanov2026mantis,feofanov2026mantisv2closingzeroshotgap} follow this protocol with transformer encoders and
tokenizers built for time series. Several of these tokenizers keep the normalized shape of
a window and its amplitude statistics in separate input
branches~\citep{lin2024nutime,feofanov2026mantis}, on the view that both are informative
for classification. MantisV2 additionally reports that test-time choices matter, in
particular which intermediate layer is read and how output tokens are
aggregated~\citep{feofanov2026mantisv2closingzeroshotgap}, and UTICA keeps the same backbone while replacing
the contrastive objective with self-distillation from a slowly updated copy of the
student~\citep{moakher2026utica}. A related body of work obtains competitive frozen
features from models never trained for classification, such as vision
transformers~\citep{roschmann2025tivit} and
forecasters~\citep{auer2025tirexclassification}, and reports the same sensitivity to layer
selection and token aggregation.

\textbf{Synthetic corpora and distillation.}
Since large curated collections of real series are scarce, pre-training data for these
models is increasingly synthetic. CauKer generates series by composing Gaussian process
kernels inside structural causal models and reports regular scaling in both corpus size
and model capacity~\citep{xie2026cauker}. Other priors have been proposed for
classification~\citep{orourke2026causaldag} and forecasting~\citep{moroshan2025tempopfn}, and mixing several generators is reported to
be preferable to selecting one~\citep{nagpal2026mix}. Distillation has been applied to time
series models mostly for compression, transferring a large pre-trained encoder into a
smaller one. Using it to merge several specialized models into a single one is a classical
idea from the distillation literature~\citep{hinton2015distilling}, and has been studied
more extensively in vision, where teachers trained under different objectives may disagree
and have to be reconciled~\citep{comad2025}.

\section{Per-dataset baseline reproduction}
\label{app:perdataset}
% \tzh{"refinement required, too many details rn"}

Our central comparison is against TS2Vec trained the way it was designed to be trained,
so the credibility of that baseline matters as much as the credibility of our own model.
We therefore validated it twice: once at the level of the per-dataset protocol we use for
the paired comparison (\S\ref{app:protocol}--\ref{app:paired}), and once by reproducing
the original paper end to end across all three of its tasks (\S\ref{app:fullrepro}).

\subsection{Protocol}
\label{app:protocol}

For each of the 128 UCR datasets we train one TS2Vec encoder from scratch with the reference implementation and its defaults: ten
dilated residual blocks, hidden width 64, a 320-dimensional representation, learning rate
$10^{-3}$, batch size 8, maximum training length 3000, Bernoulli(0.5) timestamp masking,
and the upstream iteration rule of 200 iterations for training sets with at most
$10^5$ values and 600 iterations otherwise. Frozen representations are then probed two ways: with the upstream grid-searched RBF-SVM, which
is what the original paper reports, and with the StandardScaler plus logistic regression
probe of Table~\ref{tab:main}, so that the comparison against our encoder is
probe-matched. Aggregated over all 128 datasets this gives 82.10\% (RBF-SVM) and 81.36\%
(logistic regression); the same frozen representations under the random-forest probe of
Table~\ref{tab:main} (200 trees, three predictor seeds) give 79.83\%.

The UEA-27 entries of Table~\ref{tab:main} follow the same recipe with upstream's
multivariate handling: one encoder per dataset whose input projection maps all $C$
channels jointly, series standardized per channel with training-split statistics, the
same optimizer, iteration rule and final-checkpoint readout, and the paper's two probes on
the resulting 320-dimensional embeddings. This gives 69.50\% (random forest) and 70.79\%
(logistic regression); per dataset the numbers track the published TS2Vec UEA table
closely (e.g.\ 0.99 vs.\ 0.987 on ArticularyWordRecognition, 0.32 vs.\ 0.308 on
EthanolConcentration, 0.68 vs.\ 0.682 on PEMS-SF).

\subsection{Agreement with the published numbers}
\label{app:agreement}

The original paper tabulates per-dataset accuracies for all 128 UCR datasets (its Table~9,
default batch size 8); its headline 0.830 averages the 125 datasets on which every
baseline runs, dropping the three DodgerLoop datasets with missing values. Over all 128
datasets our reproduction reaches \textbf{82.10\% against the published 82.87\%}, a gap
of 0.77 pp, with a mean absolute per-dataset deviation of \textbf{1.65 pp} and a
per-dataset Pearson correlation of \textbf{0.984}; on the paper's 125-dataset subset the
aggregates are 82.25\% against 82.96\%. Agreement is not concentrated in the aggregate:
69 of the 128 datasets are within 1 pp, 98 within 2 pp, and 119 within 5 pp.

Eight datasets deviate by more than 5 pp, all downward.
We rerun these datasets at four additional pre-training seeds
(Table~\ref{tab:perdataset}).
The reruns show substantial seed sensitivity on several of the largest discrepancies:
replacing the seed-42 values by five-seed means raises the UCR-128 aggregate from
82.10\% to 82.23\%.
PLAID remains an unresolved exception, with $34.9\pm10.1$\% across five seeds against
56.1\% published.
We retain seed 42 for the main per-dataset reproduction to keep this baseline protocol uniform.

\begin{table}[h]
  \caption{The eight datasets on which the seed-42 reproduction deviates from the published
  value by more than 5 pp, with a four-extra-seed diagnostic (per-dataset TS2Vec, RBF-SVM
  probe, \%). The main per-dataset baseline uses seed 42.}
  \label{tab:perdataset}
  \centering
  \begin{tabular}{lccccccc}
    \toprule
    & & \multicolumn{5}{c}{ours, per pre-training seed} & \\
    \cmidrule(lr){3-7}
    Dataset & published & 42 & 43 & 44 & 45 & 46 & mean $\pm$ sd \\
    \midrule
    PLAID                       & 56.1 & 37.4 & 30.9 & 19.4 & 42.8 & 44.1 & 34.9 $\pm$ 10.1 \\
    Car                         & 83.3 & 70.0 & 78.3 & 71.7 & 83.3 & 70.0 & 74.7 $\pm$ 5.9 \\
    DodgerLoopDay               & 56.2 & 47.5 & 48.8 & 48.8 & 50.0 & 52.5 & 49.5 $\pm$ 1.9 \\
    Herring                     & 64.1 & 56.2 & 59.4 & 59.4 & 68.8 & 57.8 & 60.3 $\pm$ 4.9 \\
    FaceFour                    & 93.2 & 86.4 & 84.1 & 90.9 & 89.8 & 89.8 & 88.2 $\pm$ 2.9 \\
    MiddlePhalanxOutlineCorrect & 83.8 & 77.3 & 78.0 & 83.8 & 80.1 & 79.4 & 79.7 $\pm$ 2.5 \\
    PhalangesOutlinesCorrect    & 80.9 & 75.3 & 81.0 & 81.5 & 82.1 & 79.7 & 79.9 $\pm$ 2.7 \\
    WormsTwoClass               & 80.5 & 75.3 & 75.3 & 67.5 & 76.6 & 79.2 & 74.8 $\pm$ 4.4 \\
    \bottomrule
  \end{tabular}
\end{table}

\subsection{The paired comparison}
\label{app:paired}

Table~\ref{tab:paired} gives the three comparisons summarised in
Section~\ref{sec:experiments}. Under the matched logistic-regression probe our single
frozen encoder improves on 128 dataset-specific encoders by 1.95 pp, winning on 88
datasets, tying on 9 and losing on 31; the Wilcoxon signed-rank test rejects equality at
$p=1.3\times10^{-6}$. Giving the baseline its own RBF-SVM probe, which is worth 0.74 pp
to it, leaves a 1.22 pp advantage at $p=0.02$. Against the \emph{published} per-dataset
numbers on all 128 datasets the two are statistically indistinguishable (83.32\% vs
82.87\%, $p=0.43$), which is the honest reading: our encoder clearly beats a faithful
reimplementation of per-dataset TS2Vec, and ties the numbers as originally published.

\begin{table}[h]
  \caption{TS2Vec-FM (one encoder, frozen, synthetic pre-training only) against TS2Vec
  trained per dataset (TS2Vec/ds) on UCR-128. W\,/\,T\,/\,L counts the datasets on which
  TS2Vec-FM wins, ties and loses against that row; $p$-values (two-sided Wilcoxon
  signed-rank) are given in the text.}
  \label{tab:paired}
  \centering
  \begin{tabular}{llcc}
    \toprule
    Method & Probe & UCR acc. & W\,/\,T\,/\,L \\
    & & &  of TS2Vec-FM \\
    \midrule
    TS2Vec-FM                 & LogReg  & \textbf{83.32} & --- \\
    \midrule
    TS2Vec/ds (reproduced)    & LogReg  & 81.36 & 88\,/\,9\,/\,31  \\
    TS2Vec/ds (reproduced)    & RBF-SVM & 82.10 & 71\,/\,11\,/\,46 \\
    TS2Vec/ds (published)     & RBF-SVM & 82.87 & 62\,/\,10\,/\,56 \\
    \bottomrule
  \end{tabular}
\end{table}

\subsection{Independent reproduction of the original paper}
\label{app:fullrepro}

The protocol above validates only the classification numbers we compare against. To check
that the baseline is faithful more broadly, we separately reproduced the whole TS2Vec
paper --- classification, forecasting and anomaly detection --- by running the authors'
released code \emph{unmodified} (commit \texttt{b0088e14}, no source edits) on its own
scripts and seeds. We couldn't match exactly the environment and used torch 2.5.1 with Python 3.9, i.e., bit-exact reproduction is therefore impossible by construction.

We follow \citep{yue2022ts2vec} exactly in metrics and evaluation protocol. For more details, please see their paper. Table~\ref{tab:ts2vecrepro} lists every headline metric. 15 of 16
metrics land within 5\% relative of the published value and thirteen within 3\%. The two
flagship accuracies, 0.830 on UCR-125 and 0.712 on UEA-29, come back as 0.8257 and 0.7055;
anomaly-detection precision on KPI matches to three decimals in both settings. Agreement
holds per dataset as well: 109 of 125 UCR datasets and 25 of 29 UEA datasets fall within
run-to-run noise, with mean deviations of $-0.39$ and $-0.68$ pp. Note that this table
uses the original paper's benchmark subsets, UCR-125 and UEA-29, rather than the
UCR-128 and UEA-27 used elsewhere in this paper.

Three qualifications belong with these numbers. First, the reproduction is consistently
and slightly pessimistic on classification, so the residual is best attributed to the forced library and hardware
change rather than to seed selection; 70\% of our per-dataset means still fall inside the
paper's own batch-size sensitivity band. Second, the one genuine miss is the UEA average
rank (2.776 vs 2.40). That statistic is hypersensitive on a 29-dataset benchmark where
six methods cluster tightly: a 0.9\% accuracy change produces a 15.7\% rank change, and
three datasets (DuckDuckGeese, MotorImagery, EigenWorms) contribute 8 of the 11 shifted
rank units, in each case from an accuracy change below 0.02. TS2Vec still ranks first, but
its margin over T-Loss narrows from 0.72 to 0.26 rank units. Third, coverage is
incomplete as Yahoo Webscope has been discontinued and we have not performed the experiments on the associated data. All baseline numbers are taken as published;
no baseline was re-run, so every rank above compares our TS2Vec against the paper's
reported baselines.

Taken together with \S\ref{app:agreement}, the two exercises corroborate each other. On
the paper's 125-dataset subset, with the same upstream probe and the same seed, the
per-dataset protocol of \S\ref{app:protocol} gives 82.25\% and the unmodified-code
reproduction gives 82.61\%, against a published 82.96\%. We conclude that
the baseline our main claim is measured against is a faithful TS2Vec, and that the small
residual against the published aggregate is environmental rather than methodological.

\begin{table}[h]
  \caption{Reproducing the original TS2Vec paper with unmodified released code. Accuracy
  and F1/precision/recall higher is better; MSE/MAE lower is better. $^\dagger$mean over
  three seeds (42, the paper's own, plus 43 and 44; 1--3 for anomaly detection, as in the
  released scripts). Status uses a 5\% relative tolerance.}
  \label{tab:ts2vecrepro}
  \centering
  \begin{tabular}{lllcc}
    \toprule
    Task & Benchmark & Metric & published & ours$^\dagger$ \\
    \midrule
    \multirow{6}{*}{classification}
      & UCR-125          & accuracy    & 0.830 & 0.8257  \\
      & UCR-125          & avg.\ rank  & 1.82  & 1.876   \\
      & UEA-29           & accuracy    & 0.712 & 0.7055  \\
      & UEA-29           & avg.\ rank  & 2.40  & 2.776   \\
      & UCR-125, $B{=}4$ & accuracy    & 0.824 & 0.8267  \\
      & UCR-125, $B{=}16$& accuracy    & 0.827 & 0.8250  \\
    \midrule
    \multirow{4}{*}{forecasting}
      & univariate   & avg.\ MSE & 0.209 & 0.2056  \\
      & univariate   & avg.\ MAE & 0.296 & 0.2936  \\
      & multivariate & avg.\ MSE & 0.828 & 0.8441  \\
      & multivariate & avg.\ MAE & 0.636 & 0.6420  \\
    \midrule
    \multirow{6}{*}{anomaly detection}
      & KPI, normal     & F1        & 0.677 & 0.6582  \\
      & KPI, normal     & precision & 0.929 & 0.9279  \\
      & KPI, normal     & recall    & 0.533 & 0.5110  \\
      & KPI, cold-start & F1        & 0.676 & 0.6644  \\
      & KPI, cold-start & precision & 0.907 & 0.9072  \\
      & KPI, cold-start & recall    & 0.540 & 0.5244  \\
    % \midrule
    %   & Yahoo           & F1 / P / R & \multicolumn{2}{l}{not evaluable --- dataset withdrawn} \\
    \bottomrule
  \end{tabular}
\end{table}

\section{Scaling and checkpoint selection}
\label{app:scaling}

\begin{figure}[h]
  \centering
  \includegraphics[width=\linewidth]{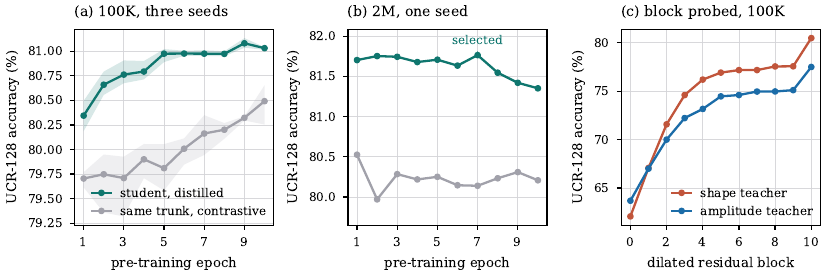}
  \caption{Colours are shared across panels; accuracy is UCR-128 with the random-forest
  probe. \textbf{(a)} At 100K, the distilled student stabilizes after the first few epochs and
  remains above the same trunk trained contrastively; bands span the three pre-training
  seeds. \textbf{(b)} At 2M (one seed) the student peaks at epoch 7 and then declines,
  while the gradient-clipped contrastive control stays within 0.56 pp of its epoch-1
  accuracy across ten epochs.
  \textbf{(c)} Probing each dilated residual block of the two 100K specialists: accuracy
  increases with depth and the final block is best for both.}
  \label{fig:curves}
\end{figure}

\textbf{Checkpoint selection.} Downstream accuracy is not monotone in pre-training epoch,
so we select the checkpoint on UCR random-forest accuracy at pre-training seed 42
(Figure~\ref{fig:curves}a,b). At 100K the student is stable: it rises to 81.13/81.07/81.04\%
at epoch 9 across the three seeds, and any epoch from 5 on is within 0.2 pp of that. At 2M
the same recipe peaks earlier and then decays --- 81.70\% at epoch 1, 81.76\% at epoch 7,
81.35\% at epoch 10 --- while the pre-training loss keeps falling, so more optimization of
the contrastive teachers' objective stops buying downstream accuracy well before
convergence. The teachers behave the same way and are selected independently: at 2M the
amplitude teacher improves over its three epochs (77.43 to 77.68\%) while the shape teacher
degrades (79.89 to 79.49\%), so they are taken at epochs 3 and 1. The specialists are
trained for three epochs only, and a 100-epoch run of both at 100K (batch 512, learning
rate $4\times10^{-3}$, seed 42) shows why: the shape specialist peaks at epoch 5 (80.65\%)
and drifts down to 79.79\% by epoch 100, the amplitude specialist peaks at epoch 10
(77.19\%) and falls to 74.67\%. Epoch 3 is within 0.4 pp of the best epoch of either
curve, and the frontier in Table~\ref{tab:frontier} shows that 16 times more updates at
batch 8 do not help either, so longer specialist training buys nothing. The contrastive control
requires a separate stability qualification. Without clipping, nominally identical 1M
runs agree through roughly 200 updates but undergo gradient explosions at different
steps and differ by 4.15 pp at the same 1954-update checkpoint. With a global gradient
clip of 10, the 1M run and its exact-seed repeat agree within 0.15 pp over their shared
three epochs. The clipped 2M control is correspondingly stable, starting at 80.52\% and
remaining between 79.97\% and 80.52\% over ten epochs.

\textbf{Input length.} A convolutional encoder accepts any length, so we evaluate at native
lengths; resampling every series to 512 steps, the preprocessing that fixed-context TSFMs
require, costs 0.20/0.22 pp on UCR and 0.90/0.56 pp on UEA with the random-forest and
logistic-regression probes (Table~\ref{tab:length}). Native lengths therefore matter more on
the multivariate archive, where series run from 25 steps (JapaneseVowels) to 17984
(EigenWorms), so one fixed context must both upsample and heavily compress, than on UCR.
The comparison matters because our baselines are evaluated at 512: the model stays
within 0.5 pp of MantisV2 on UCR under the baselines' own input handling, so the result is
not an artifact of longer inputs.

\begin{table}[h]
  \caption{Native lengths versus resampling to 512 steps, for the released 2M model
  (epoch 7, \%).}
  \label{tab:length}
  \centering
  \begin{tabular}{lcccc}
    \toprule
    Input handling & UCR RF & UCR LogReg & UEA RF & UEA LogReg \\
    \midrule
    native lengths & \textbf{81.76} & \textbf{83.32} & \textbf{73.60} & \textbf{75.30} \\
    resampled to 512 & 81.57 & 83.10 & 72.70 & 74.74 \\
    \bottomrule
  \end{tabular}
\end{table}

\textbf{Where the representation lives.} We probed each of the eleven dilated residual
blocks of both 100K specialists (Figure~\ref{fig:curves}c). Accuracy increases with depth
throughout, and the largest single jump is into the final block: 75.11 to 77.50\% for the
amplitude specialist and 77.58 to 80.49\% for the shape specialist. The ordering is
unchanged at a fixed length of 512. For both specialists the final block gives the highest UCR RF accuracy among the evaluated layers.
This comparison also changes feature dimensionality: hidden blocks output 64 channels,
whereas the final block outputs 320.
We therefore use the final block for downstream evaluation.

\textbf{Batch size and learning rate.} The teachers' optimization frontier is flat
(Table~\ref{tab:frontier}): across a 64-fold range of batch size and a fourfold range of
learning rate, the best epoch of each configuration spans 1.1 pp for the amplitude view and
0.7 pp for the shape view, with no interior optimum worth tuning for. We therefore train
the teachers at batch 128 with learning rate $10^{-3}$, which is best for the shape view and
within 0.3 pp of best for the amplitude view, and note that the reference
implementation's batch of 8 is not better despite 16 times more updates. The student is
trained at batch 512 with learning rate $4\times10^{-3}$ purely for throughput at 2M scale;
at 100K, where both are available, that substitution costs 0.18 pp (81.08 to 80.90\%).

\begin{table}[h]
  \caption{Teacher pre-training frontier at 100K (UCR random forest, best of the trained
  epochs, pre-training seed 42, \%). Dashes are configurations we did not run.}
  \label{tab:frontier}
  \centering
  \begin{tabular}{lccccccc}
    \toprule
    & \multicolumn{3}{c}{amplitude view} & & \multicolumn{3}{c}{shape view} \\
    \cmidrule(r){2-4}\cmidrule(l){6-8}
    Batch & $10^{-3}$ & $2\cdot10^{-3}$ & $4\cdot10^{-3}$ & & $10^{-3}$ & $2\cdot10^{-3}$ & $4\cdot10^{-3}$ \\
    \midrule
    8   & 77.31 & --    & --    & & 79.80 & --    & --    \\
    128 & 77.53 & 77.36 & --    & & \textbf{80.49} & 80.33 & --    \\
    256 & 77.73 & \textbf{77.82} & --    & & 80.01 & 80.34 & --    \\
    512 & 77.52 & 77.40 & 76.70 & & 79.84 & 80.25 & 80.26 \\
    \bottomrule
  \end{tabular}
\end{table}

\section{Full results}
\label{app:additional}

This section provides the complete per-dataset results that correspond to the aggregated Table~\ref{tab:main}.
Tables~\ref{tab:perds-ucr-rf}--\ref{tab:perds-uea-lr} show UCR-128 and UEA-27 performance, each under the random-forest probe (200
trees, mean over predictor seeds 0/1/2) and the StandardScaler plus logistic-regression
probe (seed 0). Input handling
follows each model's prescribed configuration: ours, TiRex, Chronos-2, TiViT-H,
TiConvNext and Catch22+ consume native lengths, while MOMENT, NuTime and Mantis
interpolate to 512. The TS2Vec/ds column is the original TS2Vec trained on each dataset's
own training split (\S\ref{app:protocol}), so unlike every other column it is not a
zero-shot result.

{\scriptsize
\setlength{\tabcolsep}{5pt}
\begin{longtable}{lrrrrrrrrrr}
  \caption{Per-dataset accuracy (\%) on UCR-128 with the random-forest probe (mean over predictor seeds 0/1/2). TS2Vec / ds is the original TS2Vec trained on each dataset (the reference row of Table~\ref{tab:main}); TS2Vec-FM is our single frozen encoder. Best per row in bold.}\label{tab:perds-ucr-rf} \\
  \toprule
  Dataset & \rotatebox{60}{Catch22+} & \rotatebox{60}{MOMENT} & \rotatebox{60}{NuTime} & \rotatebox{60}{TiRex} & \rotatebox{60}{Chronos-2} & \rotatebox{60}{TiViT-H} & \rotatebox{60}{TiConvNext} & \rotatebox{60}{MantisV2} & \rotatebox{60}{TS2Vec / ds} & \rotatebox{60}{TS2Vec-FM} \\
  \midrule
  \endfirsthead
  \multicolumn{11}{l}{\emph{Table~\ref{tab:perds-ucr-rf} continued}} \\
  \toprule
  Dataset & \rotatebox{60}{Catch22+} & \rotatebox{60}{MOMENT} & \rotatebox{60}{NuTime} & \rotatebox{60}{TiRex} & \rotatebox{60}{Chronos-2} & \rotatebox{60}{TiViT-H} & \rotatebox{60}{TiConvNext} & \rotatebox{60}{MantisV2} & \rotatebox{60}{TS2Vec / ds} & \rotatebox{60}{TS2Vec-FM} \\
  \midrule
  \endhead
  \bottomrule
  \endfoot
  \endlastfoot
  ACSF1 & 82.3 & 82.0 & 75.0 & 85.0 & 84.7 & 85.0 & 86.7 & 82.3 & \textbf{87.7} & 86.7 \\
  Adiac & 73.4 & 78.9 & 73.6 & 78.5 & 81.1 & 70.6 & 65.5 & \textbf{84.8} & 77.0 & 80.6 \\
  AllGestureWiimoteX & 59.6 & 61.4 & 65.0 & 65.0 & 60.4 & 60.3 & 66.8 & 68.4 & \textbf{70.5} & 69.1 \\
  AllGestureWiimoteY & 63.2 & 66.2 & 63.0 & 70.5 & 64.2 & 65.1 & 68.0 & 69.1 & 72.4 & \textbf{74.7} \\
  AllGestureWiimoteZ & 54.6 & 57.7 & 62.2 & 60.6 & 62.0 & 61.3 & 60.7 & 68.8 & 68.0 & \textbf{69.7} \\
  ArrowHead & 74.1 & 79.4 & 77.3 & 78.9 & 77.9 & 76.4 & 77.1 & 80.6 & 76.6 & \textbf{81.1} \\
  Beef & 68.9 & 78.9 & 66.7 & \textbf{84.4} & 73.3 & 75.6 & 82.2 & 68.9 & 67.8 & 72.2 \\
  BeetleFly & 78.3 & 95.0 & 80.0 & 91.7 & 86.7 & 93.3 & 95.0 & \textbf{98.3} & 80.0 & 93.3 \\
  BirdChicken & 85.0 & 98.3 & 96.7 & 88.3 & 90.0 & 85.0 & \textbf{100.0} & 90.0 & 95.0 & 90.0 \\
  BME & \textbf{100.0} & 98.7 & 84.7 & 96.0 & 96.7 & 97.8 & 99.3 & 99.6 & 90.2 & 97.1 \\
  Car & 75.6 & 75.6 & 76.7 & 66.7 & 73.9 & 73.9 & 72.2 & 75.6 & \textbf{78.9} & 77.2 \\
  CBF & 97.6 & 96.8 & 97.3 & 99.1 & 99.3 & \textbf{99.8} & 99.4 & 99.5 & 98.7 & 98.4 \\
  Chinatown & \textbf{98.0} & 97.8 & 92.8 & 96.9 & 97.2 & 96.3 & 87.2 & 95.3 & 97.4 & 97.2 \\
  ChlorineConcentration & 66.8 & 69.7 & 67.1 & 70.2 & \textbf{71.0} & 70.8 & 70.7 & 70.4 & 69.6 & 68.2 \\
  CinCECGTorso & 88.7 & 68.3 & 73.1 & \textbf{89.8} & 81.4 & 81.6 & 81.0 & 80.9 & 69.6 & 72.2 \\
  Coffee & 98.8 & 92.9 & 92.9 & 98.8 & 92.9 & \textbf{100.0} & \textbf{100.0} & \textbf{100.0} & 97.6 & \textbf{100.0} \\
  Computers & 73.5 & 62.7 & \textbf{76.3} & 76.0 & 74.1 & 73.3 & 74.1 & 72.1 & 64.8 & 69.9 \\
  CricketX & 69.7 & 68.5 & 66.7 & 69.2 & 69.9 & 71.4 & 66.4 & \textbf{77.4} & 68.5 & 70.8 \\
  CricketY & 69.2 & 67.4 & 68.9 & 72.2 & 71.4 & 71.3 & 66.5 & \textbf{78.8} & 70.4 & 70.7 \\
  CricketZ & 74.3 & 74.9 & 67.8 & 72.5 & 73.0 & 76.4 & 71.2 & \textbf{81.7} & 75.0 & 75.3 \\
  Crop & \textbf{75.2} & 71.4 & 67.5 & 72.0 & 72.1 & 65.5 & 64.1 & 73.4 & 73.3 & 74.0 \\
  DiatomSizeReduction & \textbf{94.0} & 88.2 & 86.2 & 86.2 & 90.0 & 86.5 & 90.1 & 83.0 & 91.5 & 86.9 \\
  DistalPhalanxOutlineAgeGroup & 71.7 & 75.3 & 73.9 & 74.1 & \textbf{78.2} & \textbf{78.2} & 77.7 & 76.0 & 72.4 & 73.1 \\
  DistalPhalanxOutlineCorrect & 79.1 & 79.6 & 77.8 & 78.7 & \textbf{80.3} & 77.9 & 75.5 & 78.9 & 78.7 & 78.7 \\
  DistalPhalanxTW & 64.7 & 65.7 & 68.8 & 66.7 & 69.3 & 67.1 & 69.5 & \textbf{70.7} & 66.2 & 68.3 \\
  DodgerLoopDay & \textbf{64.2} & 45.8 & 54.2 & 50.4 & 49.2 & 44.2 & 53.3 & 53.3 & 58.3 & 63.6 \\
  DodgerLoopGame & 83.3 & 81.2 & 78.3 & 76.3 & 83.3 & 81.6 & 78.3 & 80.0 & 85.7 & \textbf{88.2} \\
  DodgerLoopWeekend & \textbf{98.6} & 97.6 & 95.9 & 94.9 & 96.4 & 92.8 & 90.1 & 95.2 & 96.1 & 98.4 \\
  Earthquakes & \textbf{75.1} & 74.6 & 74.3 & 74.8 & 74.6 & 74.1 & 74.6 & 74.3 & 74.8 & 74.8 \\
  ECG200 & 85.0 & \textbf{87.0} & 82.3 & 86.0 & 84.0 & 83.0 & 78.3 & 83.7 & 83.0 & 84.0 \\
  ECG5000 & 94.0 & 93.8 & 93.3 & 93.3 & 93.3 & 94.0 & 93.9 & 93.8 & 93.7 & \textbf{94.3} \\
  ECGFiveDays & 79.8 & 81.3 & 78.7 & 89.5 & 91.0 & 93.5 & 97.1 & 87.1 & \textbf{99.9} & \textbf{99.9} \\
  ElectricDevices & 74.0 & 72.6 & 70.7 & 70.4 & 74.9 & \textbf{76.2} & \textbf{76.2} & 74.2 & 70.7 & 72.5 \\
  EOGHorizontalSignal & \textbf{59.5} & 56.5 & 45.2 & 53.1 & 53.9 & 53.9 & 56.3 & 58.8 & 58.4 & 58.0 \\
  EOGVerticalSignal & \textbf{50.9} & 45.3 & 26.7 & 41.4 & 38.4 & 44.2 & 37.6 & 48.2 & 46.7 & 43.0 \\
  EthanolLevel & 38.0 & 44.0 & 34.9 & 30.6 & 43.7 & 40.5 & 34.8 & 37.9 & 32.3 & \textbf{58.8} \\
  FaceAll & \textbf{76.8} & 71.8 & 64.9 & 76.5 & 70.2 & 69.1 & 68.7 & 72.1 & 74.6 & 73.6 \\
  FaceFour & 89.8 & 76.9 & 80.3 & 79.2 & 68.2 & 64.8 & 81.8 & \textbf{91.7} & 77.3 & 91.3 \\
  FacesUCR & 85.6 & 79.6 & 71.4 & 80.0 & 76.0 & 78.1 & 76.7 & 84.4 & \textbf{89.0} & 85.1 \\
  FiftyWords & \textbf{72.6} & 71.0 & 61.4 & 63.8 & 66.9 & 64.2 & 67.4 & 71.1 & 63.2 & 66.4 \\
  Fish & 76.2 & 84.8 & 91.6 & 83.2 & 86.3 & 90.3 & 90.7 & 93.1 & 90.7 & \textbf{94.1} \\
  FordA & 91.0 & 86.1 & 89.3 & \textbf{94.1} & 91.7 & 89.7 & 90.8 & 93.0 & 92.4 & 91.5 \\
  FordB & 72.9 & 73.7 & 75.3 & \textbf{81.4} & 79.2 & 77.9 & 76.3 & 79.8 & 76.6 & 77.9 \\
  FreezerRegularTrain & \textbf{100.0} & 91.1 & 97.7 & 92.6 & 97.2 & 98.3 & 99.1 & 98.6 & 98.4 & 97.9 \\
  FreezerSmallTrain & 92.5 & 78.8 & 94.0 & 82.1 & 91.8 & 93.5 & 95.6 & 93.5 & 83.2 & \textbf{96.3} \\
  Fungi & 92.3 & \textbf{100.0} & 70.1 & 84.1 & 91.0 & 91.4 & 88.4 & 94.3 & 94.4 & 98.0 \\
  GestureMidAirD1 & 67.9 & 67.4 & 65.6 & 66.2 & 66.9 & 70.0 & \textbf{73.6} & 71.3 & 62.1 & 62.8 \\
  GestureMidAirD2 & 62.1 & 57.7 & 56.7 & 65.1 & \textbf{74.4} & 66.7 & 67.9 & 61.5 & 54.1 & 61.3 \\
  GestureMidAirD3 & 39.2 & 38.5 & 39.2 & 39.0 & 35.4 & 44.1 & \textbf{46.9} & 43.8 & 38.2 & 30.8 \\
  GesturePebbleZ1 & 87.8 & 87.8 & 89.1 & 87.0 & 88.6 & 86.6 & 82.9 & \textbf{92.1} & 89.5 & 89.5 \\
  GesturePebbleZ2 & 73.8 & 86.5 & 83.8 & 87.6 & 81.9 & 84.4 & 84.4 & \textbf{91.1} & 83.8 & 85.7 \\
  GunPoint & 94.7 & 98.7 & 94.2 & 96.4 & \textbf{99.6} & 99.3 & 96.7 & 97.8 & 95.3 & 99.3 \\
  GunPointAgeSpan & 98.8 & 95.3 & 97.2 & 98.2 & 98.5 & 98.7 & 96.7 & 98.4 & 99.2 & \textbf{99.5} \\
  GunPointMaleVersusFemale & 99.4 & 99.2 & 97.9 & 99.5 & 98.8 & 99.7 & 99.8 & \textbf{100.0} & \textbf{100.0} & \textbf{100.0} \\
  GunPointOldVersusYoung & \textbf{100.0} & 96.0 & \textbf{100.0} & 96.7 & 97.8 & 98.5 & 99.4 & 99.7 & \textbf{100.0} & \textbf{100.0} \\
  Ham & 60.6 & 73.3 & 71.1 & 71.4 & 64.1 & 65.1 & 63.5 & 67.0 & \textbf{78.7} & 73.3 \\
  HandOutlines & 90.4 & 90.9 & 89.6 & 87.6 & 91.1 & 88.7 & 88.7 & 90.4 & 91.7 & \textbf{91.8} \\
  Haptics & 48.4 & 52.4 & 47.4 & 52.7 & \textbf{54.4} & 50.8 & 51.4 & 50.1 & 48.1 & 50.4 \\
  Herring & 52.1 & 58.9 & 64.6 & 63.5 & \textbf{66.7} & 59.4 & 55.7 & 63.0 & 60.9 & 64.1 \\
  HouseTwenty & 96.9 & 93.8 & 88.5 & 96.9 & 95.8 & \textbf{97.8} & 95.0 & 95.2 & 95.0 & 73.9 \\
  InlineSkate & 38.9 & 32.2 & 32.0 & 44.9 & 41.9 & 38.5 & \textbf{45.8} & 38.5 & 39.9 & 44.0 \\
  InsectEPGRegularTrain & \textbf{100.0} & 93.0 & \textbf{100.0} & 96.9 & 95.2 & \textbf{100.0} & 98.9 & \textbf{100.0} & \textbf{100.0} & \textbf{100.0} \\
  InsectEPGSmallTrain & \textbf{100.0} & 82.5 & \textbf{100.0} & 88.9 & 87.1 & 94.5 & 95.0 & \textbf{100.0} & \textbf{100.0} & \textbf{100.0} \\
  InsectWingbeatSound & 62.9 & 62.6 & 51.7 & \textbf{65.1} & 62.7 & 52.6 & 53.7 & 60.6 & 61.8 & 63.6 \\
  ItalyPowerDemand & 95.4 & 94.8 & 87.4 & \textbf{96.2} & 95.6 & 87.8 & 92.5 & 92.5 & 91.8 & 95.3 \\
  LargeKitchenAppliances & 81.7 & 73.4 & 75.6 & 80.5 & \textbf{87.3} & 81.6 & 87.0 & 76.6 & 82.7 & 87.0 \\
  Lightning2 & 73.2 & 78.7 & 67.8 & 74.3 & 72.7 & 76.0 & 74.3 & 72.7 & \textbf{80.9} & 76.0 \\
  Lightning7 & 76.7 & 73.5 & 65.8 & 73.5 & 65.3 & 74.9 & 73.1 & 72.1 & \textbf{78.1} & 72.6 \\
  Mallat & \textbf{95.6} & 91.6 & 83.1 & 91.7 & 90.2 & 90.3 & 83.9 & 88.9 & 88.1 & 93.5 \\
  Meat & 92.2 & 92.2 & 92.8 & 86.1 & \textbf{93.3} & 83.9 & 86.7 & 91.1 & \textbf{93.3} & 92.8 \\
  MedicalImages & \textbf{77.4} & 71.4 & 71.3 & 70.9 & 72.5 & 75.0 & 73.7 & 75.2 & 71.6 & 71.5 \\
  MelbournePedestrian & \textbf{96.0} & 86.4 & 91.7 & 88.1 & 89.3 & 80.7 & 81.9 & 95.1 & 94.9 & 95.8 \\
  MiddlePhalanxOutlineAgeGroup & 59.5 & 56.5 & 60.6 & 60.0 & 55.8 & 60.4 & \textbf{63.0} & 58.0 & 59.5 & 56.5 \\
  MiddlePhalanxOutlineCorrect & 81.1 & 85.8 & 78.9 & 81.8 & \textbf{87.2} & 82.5 & 81.6 & 83.4 & 84.1 & 81.3 \\
  MiddlePhalanxTW & \textbf{58.4} & 57.8 & 52.8 & 57.4 & 57.1 & 55.2 & 56.9 & 53.2 & 55.6 & 58.0 \\
  MixedShapesRegularTrain & 93.1 & 91.7 & 93.8 & 94.7 & 94.3 & 95.1 & \textbf{95.6} & 94.7 & 89.1 & 92.7 \\
  MixedShapesSmallTrain & 88.3 & 85.1 & 90.8 & 90.9 & 90.2 & \textbf{91.9} & 91.8 & 91.6 & 82.6 & 87.5 \\
  MoteStrain & 88.2 & 89.0 & \textbf{94.8} & 91.9 & 93.3 & 85.9 & 90.5 & 93.1 & 88.9 & 89.5 \\
  NonInvasiveFetalECGThorax1 & 88.8 & 88.6 & 78.0 & 86.6 & 83.0 & 81.4 & 82.4 & 86.4 & 88.0 & \textbf{91.9} \\
  NonInvasiveFetalECGThorax2 & 90.9 & 91.0 & 81.7 & 88.8 & 86.8 & 87.5 & 86.9 & 88.7 & 91.6 & \textbf{93.8} \\
  OliveOil & 84.4 & 88.9 & 70.0 & 87.8 & 85.6 & 57.8 & 85.6 & 86.7 & 86.7 & \textbf{90.0} \\
  OSULeaf & 68.3 & 75.2 & 80.0 & 91.7 & 89.5 & 94.6 & \textbf{97.9} & 93.5 & 67.2 & 81.5 \\
  PhalangesOutlinesCorrect & 82.5 & 82.2 & 77.7 & 82.0 & 83.5 & 79.6 & 79.5 & 82.4 & \textbf{83.6} & 82.7 \\
  Phoneme & 32.2 & 29.4 & 29.1 & 37.7 & \textbf{39.4} & 35.5 & 37.4 & 36.4 & 28.4 & 30.8 \\
  PickupGestureWiimoteZ & 69.3 & 58.7 & 66.7 & 73.3 & 68.0 & 82.0 & 82.0 & 78.7 & 87.3 & \textbf{88.0} \\
  PigAirwayPressure & 23.7 & 10.7 & 34.8 & 34.6 & 33.3 & 47.4 & 57.7 & 49.0 & \textbf{59.8} & 37.3 \\
  PigArtPressure & 89.1 & 53.7 & \textbf{93.6} & 87.3 & 80.6 & 81.7 & 90.9 & 93.4 & 91.2 & 92.5 \\
  PigCVP & 51.3 & 44.1 & 82.9 & 83.5 & 67.3 & 67.9 & 71.3 & \textbf{89.7} & 75.2 & 88.6 \\
  PLAID & 87.5 & 73.9 & 77.7 & 86.8 & 85.9 & 87.1 & \textbf{89.2} & 81.9 & 29.5 & 87.5 \\
  Plane & \textbf{100.0} & 99.7 & \textbf{100.0} & \textbf{100.0} & \textbf{100.0} & \textbf{100.0} & \textbf{100.0} & \textbf{100.0} & \textbf{100.0} & \textbf{100.0} \\
  PowerCons & \textbf{99.4} & 94.1 & 93.3 & 89.8 & 94.1 & 89.1 & 90.7 & 96.5 & 97.4 & 95.9 \\
  ProximalPhalanxOutlineAgeGroup & 84.7 & 83.4 & 85.0 & 85.2 & \textbf{85.7} & 85.4 & 85.4 & 82.9 & 83.4 & 84.7 \\
  ProximalPhalanxOutlineCorrect & 86.5 & 86.0 & 83.7 & 87.7 & 87.1 & 85.1 & 83.3 & 86.7 & \textbf{90.0} & 89.0 \\
  ProximalPhalanxTW & 80.3 & 81.1 & 81.5 & 80.7 & 79.8 & 77.7 & 79.2 & \textbf{82.1} & 80.7 & 81.6 \\
  RefrigerationDevices & 54.9 & 54.9 & 53.7 & 56.8 & 55.0 & 54.7 & \textbf{59.1} & 56.4 & 57.2 & 55.6 \\
  Rock & 62.0 & 83.3 & 65.3 & \textbf{90.0} & 87.3 & 89.3 & \textbf{90.0} & 82.0 & 42.0 & 74.7 \\
  ScreenType & 52.7 & 42.8 & 50.6 & 51.6 & 48.7 & \textbf{54.5} & 54.0 & 46.0 & 44.4 & 49.3 \\
  SemgHandGenderCh2 & \textbf{92.7} & 77.8 & 85.6 & 88.2 & 89.1 & 83.9 & 87.2 & 91.4 & 91.6 & 90.2 \\
  SemgHandMovementCh2 & \textbf{86.1} & 42.5 & 67.6 & 64.9 & 62.6 & 53.7 & 59.7 & 73.1 & 77.2 & 69.1 \\
  SemgHandSubjectCh2 & \textbf{88.4} & 65.0 & 76.2 & 82.4 & 82.6 & 78.2 & 82.4 & 83.4 & 85.3 & 81.5 \\
  ShakeGestureWiimoteZ & 84.7 & 84.0 & 91.3 & 88.0 & 86.0 & 84.7 & 82.7 & \textbf{93.3} & 86.7 & 91.3 \\
  ShapeletSim & 97.0 & 95.9 & 92.0 & 94.3 & \textbf{100.0} & \textbf{100.0} & \textbf{100.0} & 95.9 & 98.1 & \textbf{100.0} \\
  ShapesAll & 82.1 & 83.2 & 83.9 & 83.6 & \textbf{88.4} & 86.8 & 84.4 & 87.0 & 86.6 & 87.3 \\
  SmallKitchenAppliances & 83.0 & 71.1 & 82.0 & \textbf{83.8} & 83.7 & 82.8 & 83.0 & 83.7 & 77.5 & 78.6 \\
  SmoothSubspace & \textbf{98.7} & 95.8 & 87.3 & 92.2 & 92.9 & 93.3 & 92.7 & 95.1 & 93.8 & 95.1 \\
  SonyAIBORobotSurface1 & 84.0 & 85.4 & 80.9 & 86.2 & 66.3 & 78.8 & 76.4 & 82.3 & 81.3 & \textbf{90.6} \\
  SonyAIBORobotSurface2 & 88.4 & 82.8 & 83.9 & 82.5 & 84.7 & 90.4 & 91.9 & \textbf{92.2} & 88.1 & 89.4 \\
  StarLightCurves & 97.0 & 97.7 & 97.9 & 97.9 & 98.0 & 97.9 & 98.0 & 98.1 & 96.9 & \textbf{98.1} \\
  Strawberry & 93.3 & 95.7 & 93.6 & 95.3 & 94.3 & 92.7 & 94.1 & 95.9 & \textbf{96.1} & 95.7 \\
  SwedishLeaf & 91.1 & 92.1 & 92.2 & 93.8 & 93.8 & 94.1 & 93.8 & \textbf{95.5} & 91.4 & 93.0 \\
  Symbols & 96.2 & 93.8 & 93.9 & 93.7 & 97.5 & 97.8 & 97.6 & 97.0 & 96.4 & \textbf{98.1} \\
  SyntheticControl & 99.2 & 95.8 & 97.2 & 98.7 & 99.3 & 99.6 & \textbf{99.8} & 99.0 & 99.3 & 99.7 \\
  ToeSegmentation1 & 85.5 & 93.1 & 84.4 & 94.7 & 92.4 & 94.2 & 86.1 & \textbf{95.5} & 93.4 & 92.7 \\
  ToeSegmentation2 & 78.5 & 84.6 & 73.8 & 90.3 & 87.7 & 85.6 & 85.1 & 88.2 & 88.2 & \textbf{91.3} \\
  Trace & \textbf{100.0} & 99.0 & 99.0 & \textbf{100.0} & \textbf{100.0} & \textbf{100.0} & \textbf{100.0} & \textbf{100.0} & \textbf{100.0} & \textbf{100.0} \\
  TwoLeadECG & 85.8 & 94.8 & 91.7 & 95.1 & 93.1 & 99.4 & 98.3 & \textbf{99.6} & 97.2 & \textbf{99.6} \\
  TwoPatterns & 99.4 & 91.6 & 85.0 & 98.1 & 90.8 & 94.7 & 96.9 & 98.2 & \textbf{99.8} & 98.8 \\
  UMD & 94.0 & 98.4 & 93.5 & 91.9 & \textbf{99.3} & 98.6 & 99.1 & 98.1 & 97.0 & 98.8 \\
  UWaveGestureLibraryAll & \textbf{94.5} & 91.6 & 88.8 & 90.7 & 92.3 & 86.4 & 87.0 & 88.5 & 86.6 & 87.3 \\
  UWaveGestureLibraryX & 80.6 & 79.4 & 81.2 & 78.2 & 80.4 & 79.5 & 79.9 & \textbf{81.3} & 77.3 & 77.6 \\
  UWaveGestureLibraryY & 72.5 & 71.4 & 73.9 & 72.4 & 73.4 & 73.9 & \textbf{76.3} & 75.7 & 69.1 & 70.6 \\
  UWaveGestureLibraryZ & 75.2 & 73.4 & 75.2 & 73.6 & 74.6 & 74.3 & 75.6 & \textbf{75.9} & 74.5 & 72.9 \\
  Wafer & \textbf{99.9} & 98.8 & 99.3 & 99.9 & 98.5 & 99.5 & \textbf{99.9} & 99.4 & 99.0 & 99.5 \\
  Wine & 63.6 & 75.3 & 71.0 & 63.0 & \textbf{87.0} & 51.2 & 66.0 & 80.9 & 76.5 & 74.7 \\
  WordSynonyms & \textbf{63.9} & 61.9 & 53.0 & 52.0 & 57.1 & 56.7 & 55.7 & 61.8 & 57.9 & 56.2 \\
  Worms & 72.3 & 73.6 & 72.3 & 80.1 & 76.6 & \textbf{83.1} & 80.1 & 75.8 & 64.9 & 82.7 \\
  WormsTwoClass & 81.8 & 80.1 & 78.4 & 84.8 & 81.4 & 83.5 & 82.3 & 81.8 & \textbf{85.7} & 83.5 \\
  Yoga & 82.9 & 83.7 & 82.1 & 76.8 & 81.6 & 80.9 & 81.4 & \textbf{84.8} & 84.2 & 83.4 \\
  \midrule
  \textbf{Mean} & 79.69 & 77.89 & 77.32 & 80.13 & 80.02 & 79.43 & 80.29 & \textbf{81.95} & 79.83 & 81.76 \\
\end{longtable}
}

{\scriptsize
\setlength{\tabcolsep}{5pt}
\begin{longtable}{lrrrrrrrrrr}
  \caption{Per-dataset accuracy (\%) on UCR-128 with the logistic-regression probe. Columns as in Table~\ref{tab:perds-ucr-rf}.}\label{tab:perds-ucr-lr} \\
  \toprule
  Dataset & \rotatebox{60}{Catch22+} & \rotatebox{60}{MOMENT} & \rotatebox{60}{NuTime} & \rotatebox{60}{TiRex} & \rotatebox{60}{Chronos-2} & \rotatebox{60}{TiViT-H} & \rotatebox{60}{TiConvNext} & \rotatebox{60}{MantisV2} & \rotatebox{60}{TS2Vec / ds} & \rotatebox{60}{TS2Vec-FM} \\
  \midrule
  \endfirsthead
  \multicolumn{11}{l}{\emph{Table~\ref{tab:perds-ucr-lr} continued}} \\
  \toprule
  Dataset & \rotatebox{60}{Catch22+} & \rotatebox{60}{MOMENT} & \rotatebox{60}{NuTime} & \rotatebox{60}{TiRex} & \rotatebox{60}{Chronos-2} & \rotatebox{60}{TiViT-H} & \rotatebox{60}{TiConvNext} & \rotatebox{60}{MantisV2} & \rotatebox{60}{TS2Vec / ds} & \rotatebox{60}{TS2Vec-FM} \\
  \midrule
  \endhead
  \bottomrule
  \endfoot
  \endlastfoot
  ACSF1 & 70.0 & 75.0 & 70.0 & 75.0 & \textbf{86.0} & 83.0 & 83.0 & 76.0 & 85.0 & 85.0 \\
  Adiac & 66.5 & 77.7 & 80.1 & 78.3 & 82.9 & 71.9 & 70.6 & \textbf{83.6} & 77.5 & 80.3 \\
  AllGestureWiimoteX & 54.9 & 66.1 & 60.7 & 70.1 & 68.4 & 67.0 & 69.4 & 73.0 & 72.0 & \textbf{74.7} \\
  AllGestureWiimoteY & 52.3 & 70.3 & 65.3 & 73.1 & 70.4 & 73.7 & 72.3 & 74.7 & 75.0 & \textbf{77.1} \\
  AllGestureWiimoteZ & 48.3 & 60.6 & 54.1 & 63.4 & 66.0 & 67.7 & 65.9 & 68.1 & 65.6 & \textbf{72.3} \\
  ArrowHead & 71.4 & 77.7 & 72.6 & 78.3 & \textbf{85.1} & 82.3 & 82.3 & \textbf{85.1} & 78.9 & 82.9 \\
  Beef & 63.3 & 80.0 & 73.3 & \textbf{93.3} & 70.0 & 76.7 & 83.3 & 73.3 & 73.3 & 76.7 \\
  BeetleFly & 70.0 & \textbf{95.0} & 70.0 & 90.0 & 80.0 & 85.0 & \textbf{95.0} & 85.0 & 80.0 & 85.0 \\
  BirdChicken & 80.0 & \textbf{100.0} & \textbf{100.0} & 90.0 & 90.0 & 90.0 & 95.0 & 90.0 & 90.0 & 90.0 \\
  BME & 99.3 & 99.3 & 96.7 & 99.3 & \textbf{100.0} & \textbf{100.0} & \textbf{100.0} & \textbf{100.0} & 96.7 & 99.3 \\
  Car & 83.3 & \textbf{90.0} & 73.3 & 80.0 & 85.0 & 81.7 & 85.0 & 81.7 & 88.3 & 88.3 \\
  CBF & 97.3 & 98.9 & 99.0 & 99.7 & \textbf{100.0} & 99.9 & 99.9 & 99.7 & 99.8 & 99.6 \\
  Chinatown & 97.7 & 98.3 & 96.2 & 98.3 & \textbf{98.5} & 96.8 & 93.9 & 96.2 & 97.7 & 98.3 \\
  ChlorineConcentration & 58.9 & 77.9 & 60.9 & \textbf{80.2} & 76.7 & 76.2 & 77.7 & 71.1 & 75.9 & 76.6 \\
  CinCECGTorso & 84.3 & 76.8 & 85.5 & \textbf{95.7} & 89.9 & 93.6 & 94.2 & 85.1 & 78.1 & 80.2 \\
  Coffee & \textbf{100.0} & 96.4 & \textbf{100.0} & \textbf{100.0} & \textbf{100.0} & \textbf{100.0} & \textbf{100.0} & \textbf{100.0} & \textbf{100.0} & \textbf{100.0} \\
  Computers & 66.8 & 61.2 & 78.8 & 80.0 & 76.8 & 76.4 & \textbf{82.8} & 70.4 & 66.0 & 70.8 \\
  CricketX & 61.0 & 74.4 & 70.5 & 70.0 & 75.6 & 75.9 & 74.9 & \textbf{78.2} & 72.8 & \textbf{78.2} \\
  CricketY & 59.5 & 70.8 & 69.0 & 75.4 & 73.6 & 78.7 & 74.6 & \textbf{81.5} & 74.6 & 78.2 \\
  CricketZ & 63.3 & 74.4 & 65.9 & 73.3 & 72.3 & 79.7 & 75.9 & 81.3 & 80.0 & \textbf{81.5} \\
  Crop & 68.7 & 72.3 & 69.0 & 70.3 & 71.6 & 68.7 & 67.4 & \textbf{73.5} & 73.4 & 72.8 \\
  DiatomSizeReduction & \textbf{96.7} & 94.1 & 91.5 & 92.8 & 94.8 & \textbf{96.7} & \textbf{96.7} & 92.8 & \textbf{96.7} & 94.1 \\
  DistalPhalanxOutlineAgeGroup & 71.2 & 66.9 & 70.5 & 74.8 & 74.1 & 71.2 & 72.7 & \textbf{77.7} & 71.2 & 73.4 \\
  DistalPhalanxOutlineCorrect & 75.4 & 75.4 & 74.3 & 75.0 & 76.4 & 75.4 & \textbf{76.8} & \textbf{76.8} & \textbf{76.8} & 74.3 \\
  DistalPhalanxTW & 61.9 & 62.6 & \textbf{68.3} & 65.5 & 65.5 & 66.9 & 65.5 & \textbf{68.3} & 65.5 & 66.9 \\
  DodgerLoopDay & 57.5 & 43.8 & 55.0 & 55.0 & 52.5 & 50.0 & 56.2 & 50.0 & 51.2 & \textbf{67.5} \\
  DodgerLoopGame & 85.5 & 84.1 & 81.9 & 79.0 & 89.1 & 84.1 & 84.1 & 86.2 & 83.3 & \textbf{91.3} \\
  DodgerLoopWeekend & 97.8 & \textbf{98.6} & 96.4 & 95.7 & 96.4 & 94.9 & 91.3 & 97.1 & 96.4 & 98.4 \\
  Earthquakes & 74.1 & 70.5 & 73.4 & 69.1 & \textbf{74.8} & 70.5 & 69.8 & 71.9 & 69.8 & 72.7 \\
  ECG200 & 85.0 & 90.0 & 85.0 & 83.0 & 84.0 & 86.0 & 86.0 & 88.0 & 87.0 & \textbf{91.0} \\
  ECG5000 & 93.9 & 93.3 & 91.9 & 93.9 & 93.4 & 93.8 & 93.1 & 93.5 & 93.7 & \textbf{94.3} \\
  ECGFiveDays & 87.0 & 97.0 & 89.9 & 97.6 & 99.1 & 97.9 & 99.0 & 99.3 & \textbf{100.0} & 99.7 \\
  ElectricDevices & 67.5 & 74.0 & 67.9 & 67.1 & 71.7 & 76.3 & \textbf{77.6} & 71.8 & 66.4 & 68.4 \\
  EOGHorizontalSignal & 52.8 & 56.4 & 53.3 & 57.2 & 58.8 & 59.7 & 57.5 & 56.9 & 54.1 & \textbf{60.8} \\
  EOGVerticalSignal & 47.8 & 50.0 & 33.1 & 41.7 & 43.9 & 47.2 & 42.5 & \textbf{51.4} & 49.7 & 42.5 \\
  EthanolLevel & 56.8 & \textbf{68.2} & 52.0 & 50.8 & 65.4 & 58.4 & 58.0 & 59.2 & 48.6 & 64.6 \\
  FaceAll & 80.2 & 77.9 & 72.1 & 78.8 & 75.3 & 75.2 & 73.7 & 75.7 & \textbf{89.3} & 79.4 \\
  FaceFour & 88.6 & 87.5 & 96.6 & 95.5 & 90.9 & 87.5 & 92.0 & \textbf{98.9} & 89.8 & 94.3 \\
  FacesUCR & 85.0 & 89.1 & 82.0 & 87.8 & 85.8 & 87.5 & 85.6 & 91.2 & \textbf{94.1} & 93.8 \\
  FiftyWords & 72.3 & 77.6 & 71.6 & 72.3 & 79.6 & 78.7 & 77.1 & \textbf{82.0} & 79.8 & \textbf{82.0} \\
  Fish & 86.3 & 96.0 & 91.4 & 90.3 & 94.3 & 94.9 & 97.1 & 96.0 & \textbf{97.7} & 96.6 \\
  FordA & 87.4 & 90.6 & 90.6 & \textbf{94.7} & 92.8 & 91.4 & 92.3 & 94.3 & 90.7 & 91.2 \\
  FordB & 72.7 & 75.6 & 76.5 & \textbf{81.6} & 80.7 & 79.6 & 77.9 & 81.0 & 76.4 & 73.6 \\
  FreezerRegularTrain & 99.2 & 98.8 & 99.4 & 99.1 & 99.5 & 99.6 & \textbf{99.8} & 99.6 & 99.1 & 99.2 \\
  FreezerSmallTrain & 94.1 & 81.9 & 98.6 & 89.1 & 96.9 & 98.4 & \textbf{99.0} & 98.5 & 93.5 & 94.2 \\
  Fungi & 91.9 & \textbf{100.0} & 74.7 & 95.2 & 94.6 & 97.8 & 99.5 & 97.3 & 96.2 & 97.8 \\
  GestureMidAirD1 & 62.3 & 68.5 & 68.5 & 69.2 & 73.1 & 75.4 & 76.2 & \textbf{77.7} & 66.2 & 66.2 \\
  GestureMidAirD2 & 63.8 & 57.7 & 56.2 & 61.5 & \textbf{70.0} & 66.2 & 67.7 & 64.6 & 60.8 & 68.5 \\
  GestureMidAirD3 & 40.8 & 36.9 & 44.6 & 40.8 & 35.4 & 46.9 & \textbf{49.2} & 44.6 & 39.2 & 35.4 \\
  GesturePebbleZ1 & 82.0 & 89.5 & 89.5 & 88.4 & 90.1 & 90.7 & 85.5 & \textbf{93.0} & 91.9 & 90.7 \\
  GesturePebbleZ2 & 70.3 & 89.2 & 77.2 & 84.2 & 86.7 & 85.4 & 81.6 & \textbf{90.5} & 86.1 & 84.2 \\
  GunPoint & 92.7 & \textbf{100.0} & 97.3 & 98.0 & 99.3 & 99.3 & 99.3 & 98.0 & 98.0 & \textbf{100.0} \\
  GunPointAgeSpan & 94.3 & 97.8 & 96.8 & 99.1 & 99.1 & \textbf{99.4} & 98.7 & \textbf{99.4} & 98.7 & 98.1 \\
  GunPointMaleVersusFemale & \textbf{100.0} & 99.7 & 98.7 & 99.4 & 99.4 & \textbf{100.0} & 99.7 & \textbf{100.0} & \textbf{100.0} & \textbf{100.0} \\
  GunPointOldVersusYoung & \textbf{100.0} & 92.7 & \textbf{100.0} & 96.5 & 97.8 & 98.7 & 99.4 & 99.7 & \textbf{100.0} & \textbf{100.0} \\
  Ham & 62.9 & 65.7 & 73.3 & 70.5 & 69.5 & \textbf{74.3} & 72.4 & 70.5 & 73.3 & 73.3 \\
  HandOutlines & 88.4 & 94.1 & 88.6 & 87.6 & 93.8 & \textbf{94.3} & 92.2 & 93.0 & 91.9 & 93.8 \\
  Haptics & 41.9 & 49.0 & 42.5 & 47.7 & 50.3 & 49.4 & \textbf{52.3} & 50.3 & 48.4 & 49.7 \\
  Herring & 57.8 & 64.1 & 53.1 & 60.9 & 51.6 & 53.1 & 60.9 & 68.8 & 67.2 & \textbf{70.3} \\
  HouseTwenty & 95.0 & 93.3 & 89.9 & \textbf{98.3} & 97.5 & \textbf{98.3} & \textbf{98.3} & 93.3 & 93.3 & 78.2 \\
  InlineSkate & 37.8 & 36.4 & 35.6 & 43.8 & 43.6 & 39.8 & \textbf{44.7} & 39.1 & 42.4 & 44.5 \\
  InsectEPGRegularTrain & 99.2 & 94.8 & \textbf{100.0} & 99.6 & \textbf{100.0} & \textbf{100.0} & \textbf{100.0} & \textbf{100.0} & \textbf{100.0} & \textbf{100.0} \\
  InsectEPGSmallTrain & 93.2 & 83.5 & \textbf{100.0} & 95.6 & 92.0 & 98.8 & 98.8 & \textbf{100.0} & \textbf{100.0} & \textbf{100.0} \\
  InsectWingbeatSound & 58.2 & 62.1 & 51.0 & 62.5 & 63.1 & 55.1 & 56.5 & 59.6 & \textbf{63.4} & 62.6 \\
  ItalyPowerDemand & 93.9 & 95.4 & 91.2 & 96.0 & \textbf{96.4} & 94.1 & 93.8 & 94.6 & 95.5 & 96.2 \\
  LargeKitchenAppliances & 83.7 & 76.0 & 77.1 & 82.9 & 86.1 & 87.2 & \textbf{89.3} & 81.6 & 85.6 & \textbf{89.3} \\
  Lightning2 & 75.4 & 78.7 & 73.8 & 73.8 & 78.7 & 82.0 & 83.6 & 78.7 & \textbf{88.5} & 80.3 \\
  Lightning7 & 69.9 & 74.0 & 78.1 & 80.8 & 79.5 & 84.9 & 76.7 & \textbf{86.3} & 84.9 & 78.1 \\
  Mallat & 88.7 & 91.9 & 84.9 & 95.5 & 94.7 & \textbf{97.7} & 96.4 & 92.0 & 88.5 & 91.1 \\
  Meat & 93.3 & \textbf{100.0} & 93.3 & 90.0 & 93.3 & 81.7 & 85.0 & 96.7 & 91.7 & 93.3 \\
  MedicalImages & 67.9 & 74.6 & 70.9 & 72.5 & 74.7 & 73.2 & 75.1 & 77.1 & 76.1 & \textbf{78.7} \\
  MelbournePedestrian & 90.3 & 89.5 & 93.3 & 89.4 & 90.5 & 86.3 & 87.0 & 95.8 & 94.7 & \textbf{96.5} \\
  MiddlePhalanxOutlineAgeGroup & \textbf{55.8} & 50.6 & 55.2 & 48.7 & 55.2 & \textbf{55.8} & 52.6 & 54.5 & 50.6 & 48.1 \\
  MiddlePhalanxOutlineCorrect & 77.7 & 80.8 & 74.9 & 80.8 & 83.5 & 79.4 & 81.8 & \textbf{84.2} & 78.7 & 82.8 \\
  MiddlePhalanxTW & 54.5 & 54.5 & 44.2 & 50.0 & 50.0 & 51.9 & 48.1 & 51.3 & 51.3 & \textbf{55.2} \\
  MixedShapesRegularTrain & 91.5 & 93.6 & 93.6 & 97.0 & 96.7 & \textbf{97.8} & \textbf{97.8} & 97.4 & 91.5 & 94.6 \\
  MixedShapesSmallTrain & 86.9 & 89.0 & 92.1 & 93.6 & 93.9 & 95.0 & \textbf{95.4} & 94.6 & 85.2 & 91.3 \\
  MoteStrain & 85.0 & 90.5 & \textbf{95.5} & 92.4 & 92.6 & 91.6 & 90.7 & 93.9 & 87.4 & 89.3 \\
  NonInvasiveFetalECGThorax1 & 89.9 & 92.4 & 85.4 & 89.9 & 88.4 & 90.0 & 89.1 & 92.9 & 93.1 & \textbf{95.1} \\
  NonInvasiveFetalECGThorax2 & 91.8 & 93.7 & 90.3 & 90.3 & 91.3 & 92.4 & 92.3 & 93.6 & 93.7 & \textbf{95.4} \\
  OliveOil & 83.3 & 90.0 & 76.7 & \textbf{93.3} & 83.3 & 73.3 & 90.0 & 86.7 & 83.3 & 90.0 \\
  OSULeaf & 73.6 & 79.3 & 86.0 & 93.8 & 93.8 & 98.3 & \textbf{99.2} & 96.3 & 77.7 & 86.8 \\
  PhalangesOutlinesCorrect & 74.6 & 81.4 & 74.7 & 79.3 & \textbf{82.3} & 80.0 & 78.3 & 79.5 & 80.8 & 81.4 \\
  Phoneme & 24.2 & 30.0 & 27.6 & 39.0 & \textbf{41.7} & 39.6 & 39.7 & 35.5 & 28.1 & 30.4 \\
  PickupGestureWiimoteZ & 66.0 & 70.0 & 64.0 & 84.0 & 66.0 & \textbf{92.0} & 88.0 & 76.0 & 88.0 & 90.0 \\
  PigAirwayPressure & 14.4 & 11.1 & 45.2 & 41.8 & 41.8 & 63.0 & \textbf{74.5} & 57.7 & 56.2 & 32.7 \\
  PigArtPressure & 76.4 & 54.8 & 94.2 & 88.9 & 84.1 & 88.9 & 93.3 & 90.9 & \textbf{94.7} & 88.9 \\
  PigCVP & 41.8 & 53.4 & 88.5 & 86.5 & 78.8 & 79.3 & 83.2 & 93.8 & 84.1 & \textbf{94.2} \\
  PLAID & 65.4 & 78.6 & 75.6 & 87.0 & 88.6 & 91.8 & \textbf{93.1} & 85.3 & 26.3 & 90.5 \\
  Plane & 99.0 & \textbf{100.0} & \textbf{100.0} & \textbf{100.0} & \textbf{100.0} & \textbf{100.0} & \textbf{100.0} & \textbf{100.0} & \textbf{100.0} & \textbf{100.0} \\
  PowerCons & \textbf{98.9} & 95.6 & 97.2 & 93.9 & 93.3 & 90.6 & 92.2 & 97.2 & 96.7 & 97.2 \\
  ProximalPhalanxOutlineAgeGroup & 83.9 & 82.9 & \textbf{85.4} & 84.9 & 83.4 & 82.9 & 82.9 & 83.9 & 81.5 & 82.4 \\
  ProximalPhalanxOutlineCorrect & 84.5 & 88.7 & 83.8 & 88.7 & 88.3 & 87.3 & 86.3 & 87.6 & 88.3 & \textbf{90.4} \\
  ProximalPhalanxTW & 79.0 & \textbf{79.5} & 78.5 & 76.1 & 75.6 & 76.1 & 74.6 & 76.1 & 78.5 & \textbf{79.5} \\
  RefrigerationDevices & 53.3 & 51.5 & 56.3 & 54.1 & 56.8 & 57.6 & \textbf{59.7} & 52.5 & 57.9 & 57.9 \\
  Rock & 80.0 & 90.0 & 76.0 & 88.0 & 90.0 & \textbf{96.0} & 92.0 & 84.0 & 64.0 & 90.0 \\
  ScreenType & 46.9 & 39.2 & 46.9 & 52.8 & \textbf{53.9} & 53.6 & 49.9 & 50.9 & 38.7 & 46.9 \\
  SemgHandGenderCh2 & \textbf{92.0} & 76.5 & 86.2 & 85.8 & 88.8 & 90.2 & 91.0 & 89.7 & 90.5 & 90.0 \\
  SemgHandMovementCh2 & 65.8 & 44.4 & 62.4 & 52.0 & 57.6 & 56.0 & 59.1 & 62.4 & \textbf{80.9} & 62.4 \\
  SemgHandSubjectCh2 & 82.7 & 70.9 & 75.6 & 81.3 & 84.4 & 86.4 & 84.9 & 85.3 & \textbf{90.4} & 85.3 \\
  ShakeGestureWiimoteZ & 82.0 & 84.0 & 90.0 & 88.0 & 92.0 & 84.0 & 86.0 & 92.0 & \textbf{94.0} & 92.0 \\
  ShapeletSim & 95.6 & 96.7 & 92.8 & 96.7 & \textbf{100.0} & \textbf{100.0} & \textbf{100.0} & 97.8 & 97.8 & \textbf{100.0} \\
  ShapesAll & 78.0 & 84.8 & 87.2 & 86.3 & 89.7 & \textbf{91.2} & 89.7 & 90.2 & 88.2 & 91.0 \\
  SmallKitchenAppliances & 77.3 & 61.1 & \textbf{82.9} & 81.3 & 81.1 & 82.1 & 82.7 & 78.7 & 70.9 & 72.5 \\
  SmoothSubspace & 94.0 & \textbf{98.0} & 94.7 & 96.7 & 96.7 & 96.7 & \textbf{98.0} & 97.3 & 96.7 & \textbf{98.0} \\
  SonyAIBORobotSurface1 & 75.9 & 82.5 & 81.5 & \textbf{93.7} & 82.4 & 86.4 & 85.0 & 83.4 & 82.9 & 89.5 \\
  SonyAIBORobotSurface2 & 92.8 & 89.1 & 88.4 & 89.1 & 88.5 & 93.8 & \textbf{94.5} & 93.7 & 87.2 & 93.5 \\
  StarLightCurves & 96.0 & 96.4 & 97.5 & 97.4 & 97.5 & 97.4 & 97.2 & 97.3 & 96.2 & \textbf{97.6} \\
  Strawberry & 94.3 & 96.2 & 95.9 & 96.5 & 97.3 & 95.7 & 95.4 & 96.2 & 96.2 & \textbf{98.1} \\
  SwedishLeaf & 89.8 & 93.1 & 93.9 & 95.4 & 94.7 & 95.8 & 94.6 & \textbf{97.3} & 93.1 & 94.7 \\
  Symbols & 95.5 & 95.6 & 96.9 & 96.7 & 98.4 & \textbf{98.6} & \textbf{98.6} & 97.7 & 95.6 & 92.4 \\
  SyntheticControl & 98.0 & 97.7 & 97.3 & 99.3 & 99.7 & \textbf{100.0} & \textbf{100.0} & 99.0 & 99.7 & 99.3 \\
  ToeSegmentation1 & 75.9 & 94.7 & 84.2 & \textbf{96.9} & 91.7 & 93.9 & 90.4 & 96.5 & 93.4 & 93.9 \\
  ToeSegmentation2 & 76.9 & 90.8 & 86.9 & 91.5 & 86.2 & 90.0 & 87.7 & \textbf{93.1} & 92.3 & 91.5 \\
  Trace & \textbf{100.0} & 99.0 & \textbf{100.0} & \textbf{100.0} & \textbf{100.0} & \textbf{100.0} & \textbf{100.0} & \textbf{100.0} & \textbf{100.0} & \textbf{100.0} \\
  TwoLeadECG & 95.1 & 99.2 & 96.2 & 98.5 & 99.3 & \textbf{100.0} & 99.3 & 99.9 & 99.5 & 99.8 \\
  TwoPatterns & 97.9 & 99.2 & 94.2 & 99.6 & 99.5 & 99.8 & 99.9 & 99.7 & \textbf{100.0} & 99.9 \\
  UMD & \textbf{99.3} & \textbf{99.3} & 97.2 & 95.8 & \textbf{99.3} & \textbf{99.3} & \textbf{99.3} & \textbf{99.3} & 98.6 & \textbf{99.3} \\
  UWaveGestureLibraryAll & 91.7 & \textbf{96.0} & 93.2 & 94.3 & 95.9 & 94.4 & 93.7 & 95.1 & 92.0 & 93.0 \\
  UWaveGestureLibraryX & 78.1 & 79.3 & 80.0 & 80.0 & 82.1 & 83.1 & 84.2 & \textbf{84.5} & 78.2 & 81.2 \\
  UWaveGestureLibraryY & 68.8 & 71.4 & 69.6 & 70.7 & 73.9 & \textbf{77.4} & \textbf{77.4} & 76.6 & 66.6 & 72.9 \\
  UWaveGestureLibraryZ & 71.7 & 74.0 & 75.5 & 73.3 & 75.5 & 77.4 & \textbf{77.4} & \textbf{77.5} & 72.3 & 74.9 \\
  Wafer & 99.4 & 99.4 & 99.6 & 99.9 & 99.8 & 99.9 & \textbf{100.0} & 99.7 & 99.6 & 99.6 \\
  Wine & 70.4 & 83.3 & 77.8 & \textbf{87.0} & 85.2 & 66.7 & 83.3 & 83.3 & 79.6 & 77.8 \\
  WordSynonyms & 56.4 & 68.5 & 60.0 & 59.9 & 66.8 & 67.7 & 67.7 & \textbf{71.2} & 68.3 & 71.0 \\
  Worms & 58.4 & 68.8 & 76.6 & 79.2 & 77.9 & 77.9 & \textbf{85.7} & 77.9 & 64.9 & 79.2 \\
  WormsTwoClass & 72.7 & 75.3 & 67.5 & 80.5 & 80.5 & 81.8 & \textbf{84.4} & 79.2 & 63.6 & 77.9 \\
  Yoga & 71.1 & 85.1 & 74.9 & 79.5 & 81.9 & 84.0 & \textbf{87.9} & 85.6 & 85.0 & 86.2 \\
  \midrule
  \textbf{Mean} & 77.04 & 79.84 & 79.05 & 82.02 & 82.44 & 82.88 & 83.47 & \textbf{83.60} & 81.36 & 83.32 \\
\end{longtable}
}

{\scriptsize
\setlength{\tabcolsep}{5pt}
\begin{longtable}{lrrrrrrrrrr}
  \caption{Per-dataset accuracy (\%) on UEA-27 with the random-forest probe (mean over predictor seeds 0/1/2). TS2Vec/ds is the original TS2Vec trained on each dataset (the reference row of Table~\ref{tab:main}); TS2Vec-FM is our single frozen encoder. Best per row in bold.}\label{tab:perds-uea-rf} \\
  \toprule
  Dataset & \rotatebox{60}{Catch22+} & \rotatebox{60}{MOMENT} & \rotatebox{60}{NuTime} & \rotatebox{60}{TiRex} & \rotatebox{60}{Chronos-2} & \rotatebox{60}{TiViT-H} & \rotatebox{60}{TiConvNext} & \rotatebox{60}{MantisV2} & \rotatebox{60}{TS2Vec/ds} & \rotatebox{60}{TS2Vec-FM} \\
  \midrule
  \endfirsthead
  \multicolumn{11}{l}{\emph{Table~\ref{tab:perds-uea-rf} continued}} \\
  \toprule
  Dataset & \rotatebox{60}{Catch22+} & \rotatebox{60}{MOMENT} & \rotatebox{60}{NuTime} & \rotatebox{60}{TiRex} & \rotatebox{60}{Chronos-2} & \rotatebox{60}{TiViT-H} & \rotatebox{60}{TiConvNext} & \rotatebox{60}{MantisV2} & \rotatebox{60}{TS2Vec/ds} & \rotatebox{60}{TS2Vec-FM} \\
  \midrule
  \endhead
  \bottomrule
  \endfoot
  \endlastfoot
  ArticularyWordRecognition & 96.0 & 98.4 & 99.1 & 99.0 & \textbf{99.3} & 98.3 & 98.0 & 99.2 & 96.0 & 99.1 \\
  BasicMotions & \textbf{100.0} & 97.5 & \textbf{100.0} & 96.7 & \textbf{100.0} & 98.3 & \textbf{100.0} & \textbf{100.0} & 99.2 & \textbf{100.0} \\
  CharacterTrajectories & 98.1 & 97.0 & 96.7 & 96.0 & 96.3 & 93.7 & 95.2 & 97.6 & \textbf{98.7} & 98.6 \\
  Cricket & 94.4 & 97.2 & 99.1 & 99.5 & 98.6 & \textbf{100.0} & 99.5 & 98.6 & 97.7 & 98.6 \\
  DuckDuckGeese & 51.3 & 44.7 & 42.7 & 42.0 & 41.3 & 43.3 & 44.0 & \textbf{54.0} & 53.3 & 50.7 \\
  EigenWorms & 87.5 & 76.1 & 73.5 & 78.6 & 80.2 & 88.8 & \textbf{93.1} & 81.4 & 82.4 & 88.0 \\
  Epilepsy & 98.6 & 99.3 & \textbf{100.0} & \textbf{100.0} & \textbf{100.0} & \textbf{100.0} & \textbf{100.0} & 99.8 & 92.5 & 98.8 \\
  ERing & 88.3 & 96.8 & 97.7 & 97.7 & \textbf{98.8} & 98.1 & 96.3 & \textbf{98.8} & 84.2 & 92.7 \\
  EthanolConcentration & \textbf{42.7} & 29.5 & 41.3 & 27.5 & 34.9 & 38.7 & 36.6 & 41.8 & 30.9 & 35.6 \\
  FaceDetection & 52.7 & 55.4 & 56.9 & \textbf{61.3} & 57.2 & 55.0 & 53.2 & 54.9 & 50.9 & 53.5 \\
  FingerMovements & 49.7 & 53.0 & 52.7 & 52.0 & 52.3 & 53.7 & 53.3 & \textbf{55.0} & 52.0 & 54.0 \\
  HandMovementDirection & \textbf{33.8} & 29.3 & 28.8 & 31.5 & 27.5 & 28.8 & 32.9 & 27.9 & 32.9 & 29.3 \\
  Handwriting & 28.1 & 26.0 & 20.5 & 25.3 & 25.8 & 23.7 & 24.9 & 28.1 & \textbf{39.6} & 30.7 \\
  Heartbeat & 75.0 & 73.2 & 77.6 & 72.5 & 73.2 & 72.5 & 73.5 & \textbf{79.2} & 74.8 & 77.6 \\
  InsectWingbeatSubset & 36.2 & 26.7 & 61.4 & 28.7 & 28.0 & 32.0 & 32.4 & 60.7 & 27.4 & \textbf{63.9} \\
  JapaneseVowels & 95.5 & 88.5 & 93.4 & 89.3 & 81.6 & 88.2 & 87.2 & 94.0 & \textbf{98.2} & 94.0 \\
  Libras & 83.9 & 84.6 & 87.8 & 89.6 & 86.5 & 90.4 & 90.7 & \textbf{92.4} & 82.8 & 83.3 \\
  LSST & 61.6 & 63.1 & 56.7 & 57.3 & 58.9 & 59.5 & 60.0 & 66.0 & 54.3 & \textbf{66.7} \\
  MotorImagery & 43.3 & 52.3 & 48.0 & 49.3 & 53.0 & 51.0 & 49.0 & 46.7 & 49.3 & \textbf{55.0} \\
  NATOPS & 71.5 & 83.5 & 83.3 & 83.0 & 83.9 & 86.9 & 85.7 & 88.9 & \textbf{89.3} & 89.1 \\
  PEMS-SF & 84.0 & 99.6 & 99.2 & 99.6 & 99.6 & 97.3 & 97.7 & 99.6 & 67.1 & \textbf{100.0} \\
  PhonemeSpectra & 25.0 & 21.1 & 26.6 & 26.9 & 27.1 & 27.1 & 27.1 & \textbf{32.1} & 20.3 & 25.9 \\
  RacketSports & 80.0 & 84.6 & \textbf{91.2} & 83.6 & 82.5 & 85.3 & 85.1 & 91.0 & 85.5 & 89.0 \\
  SelfRegulationSCP1 & 77.0 & 77.5 & 79.5 & 78.8 & 78.5 & 79.9 & 79.3 & \textbf{81.3} & 78.7 & 76.2 \\
  SelfRegulationSCP2 & 49.3 & 49.1 & 50.7 & 49.6 & 51.7 & 49.1 & 50.6 & 51.7 & \textbf{55.2} & 52.4 \\
  SpokenArabicDigits & 89.0 & 94.5 & 90.2 & 75.5 & 76.2 & 89.3 & 91.4 & 93.7 & 94.8 & \textbf{96.0} \\
  UWaveGestureLibrary & 87.6 & \textbf{89.2} & 88.9 & 86.1 & 88.3 & 85.4 & 81.6 & 89.0 & 88.5 & 88.5 \\
  \midrule
  \textbf{Mean} & 69.63 & 69.91 & 71.99 & 69.52 & 69.67 & 70.91 & 71.05 & \textbf{74.20} & 69.50 & 73.60 \\
\end{longtable}
}

{\scriptsize
\setlength{\tabcolsep}{5pt}
\begin{longtable}{lrrrrrrrrrr}
  \caption{Per-dataset accuracy (\%) on UEA-27 with the logistic-regression probe. Columns as in Table~\ref{tab:perds-uea-rf}.}\label{tab:perds-uea-lr} \\
  \toprule
  Dataset & \rotatebox{60}{Catch22+} & \rotatebox{60}{MOMENT} & \rotatebox{60}{NuTime} & \rotatebox{60}{TiRex} & \rotatebox{60}{Chronos-2} & \rotatebox{60}{TiViT-H} & \rotatebox{60}{TiConvNext} & \rotatebox{60}{MantisV2} & \rotatebox{60}{TS2Vec/ds} & \rotatebox{60}{TS2Vec-FM} \\
  \midrule
  \endfirsthead
  \multicolumn{11}{l}{\emph{Table~\ref{tab:perds-uea-lr} continued}} \\
  \toprule
  Dataset & \rotatebox{60}{Catch22+} & \rotatebox{60}{MOMENT} & \rotatebox{60}{NuTime} & \rotatebox{60}{TiRex} & \rotatebox{60}{Chronos-2} & \rotatebox{60}{TiViT-H} & \rotatebox{60}{TiConvNext} & \rotatebox{60}{MantisV2} & \rotatebox{60}{TS2Vec/ds} & \rotatebox{60}{TS2Vec-FM} \\
  \midrule
  \endhead
  \bottomrule
  \endfoot
  \endlastfoot
  ArticularyWordRecognition & 93.3 & 98.7 & \textbf{99.3} & 99.0 & \textbf{99.3} & 98.3 & 97.7 & \textbf{99.3} & 99.0 & \textbf{99.3} \\
  BasicMotions & \textbf{100.0} & \textbf{100.0} & \textbf{100.0} & \textbf{100.0} & \textbf{100.0} & \textbf{100.0} & \textbf{100.0} & \textbf{100.0} & \textbf{100.0} & \textbf{100.0} \\
  CharacterTrajectories & 98.4 & 98.4 & 98.5 & 96.9 & 97.4 & 96.6 & 97.1 & 98.7 & \textbf{99.4} & 99.0 \\
  Cricket & 90.3 & 98.6 & \textbf{100.0} & \textbf{100.0} & \textbf{100.0} & \textbf{100.0} & \textbf{100.0} & 98.6 & 98.6 & \textbf{100.0} \\
  DuckDuckGeese & 42.0 & 48.0 & 50.0 & 38.0 & 34.0 & 48.0 & \textbf{56.0} & 50.0 & \textbf{56.0} & \textbf{56.0} \\
  EigenWorms & 76.3 & 83.2 & 87.0 & 84.0 & 84.0 & 93.1 & \textbf{96.2} & 88.5 & 80.9 & 90.8 \\
  Epilepsy & 96.4 & 99.3 & \textbf{100.0} & \textbf{100.0} & \textbf{100.0} & \textbf{100.0} & \textbf{100.0} & \textbf{100.0} & 94.9 & 99.3 \\
  ERing & 90.7 & 97.0 & 98.1 & 97.8 & 97.8 & 98.5 & 98.1 & \textbf{98.9} & 89.3 & 96.7 \\
  EthanolConcentration & 37.3 & 47.1 & 41.1 & 40.3 & \textbf{54.8} & 50.6 & 46.0 & 42.6 & 31.6 & 37.6 \\
  FaceDetection & 54.7 & 62.1 & 61.9 & \textbf{66.0} & 63.6 & 60.0 & 58.1 & 58.3 & 50.3 & 59.6 \\
  FingerMovements & 52.0 & 57.0 & 61.0 & 62.0 & 60.0 & 57.0 & 53.0 & \textbf{64.0} & 59.0 & 56.0 \\
  HandMovementDirection & 35.1 & 31.1 & 37.8 & 32.4 & 33.8 & 29.7 & \textbf{39.2} & 31.1 & 29.7 & 32.4 \\
  Handwriting & 23.1 & 37.9 & 25.3 & 41.6 & 35.4 & 33.4 & 34.7 & 39.9 & \textbf{52.9} & 47.2 \\
  Heartbeat & 75.6 & 70.2 & 74.1 & 77.1 & 75.1 & 72.7 & 74.1 & \textbf{79.5} & 68.8 & 78.0 \\
  InsectWingbeatSubset & 32.4 & 29.8 & \textbf{61.8} & 39.3 & 34.4 & 41.1 & 42.2 & 60.4 & 31.0 & 60.9 \\
  JapaneseVowels & 96.5 & 92.4 & 98.1 & 94.1 & 90.0 & 94.9 & 94.3 & 95.9 & \textbf{98.6} & 97.6 \\
  Libras & 84.4 & 85.6 & 91.7 & 89.4 & 90.6 & 92.8 & 92.8 & \textbf{94.4} & 83.3 & 91.1 \\
  LSST & 50.9 & 60.5 & 55.7 & 59.2 & 60.1 & 60.1 & 59.1 & \textbf{65.9} & 50.7 & 64.3 \\
  MotorImagery & 50.0 & 49.0 & \textbf{59.0} & 48.0 & 53.0 & 57.0 & 50.0 & 50.0 & 50.0 & 55.0 \\
  NATOPS & 81.7 & 84.4 & 86.7 & 83.9 & 82.2 & 87.2 & 87.2 & 89.4 & \textbf{95.0} & 92.2 \\
  PEMS-SF & 71.1 & 78.6 & \textbf{89.6} & 77.5 & 81.5 & 70.5 & 80.3 & 83.2 & 67.1 & 82.1 \\
  PhonemeSpectra & 20.6 & 19.9 & 24.8 & 23.6 & 24.3 & 25.2 & 25.7 & \textbf{32.2} & 17.2 & 22.6 \\
  RacketSports & 73.0 & 84.9 & \textbf{91.4} & 88.8 & 86.8 & 86.8 & 85.5 & 90.1 & 85.5 & 89.5 \\
  SelfRegulationSCP1 & 74.4 & 82.6 & 85.7 & 85.0 & 85.0 & 85.0 & 85.0 & \textbf{86.7} & 80.2 & 81.9 \\
  SelfRegulationSCP2 & 47.8 & 51.1 & \textbf{55.0} & 49.4 & 48.3 & 52.8 & 49.4 & 51.1 & 53.9 & 54.4 \\
  SpokenArabicDigits & 90.0 & 97.9 & 96.5 & 96.4 & 91.6 & 96.0 & 96.7 & 98.0 & 98.1 & \textbf{98.8} \\
  UWaveGestureLibrary & 85.9 & 91.6 & \textbf{92.5} & 87.2 & \textbf{92.5} & 91.9 & 85.9 & 92.2 & 90.3 & 90.6 \\
  \midrule
  \textbf{Mean} & 67.56 & 71.74 & 74.91 & 72.48 & 72.43 & 73.30 & 73.50 & \textbf{75.52} & 70.79 & 75.30 \\
\end{longtable}
}

\end{document}